\pdfoutput=1
\documentclass[11pt]{article}

\usepackage[preprint]{acl}

\usepackage{times}
\usepackage{latexsym}
\usepackage{comment}
\usepackage[T1]{fontenc}
\usepackage[utf8]{inputenc}

\usepackage{microtype}

\usepackage{inconsolata}

\usepackage{graphicx}

\usepackage{amsmath}
\usepackage{autobreak}

\usepackage{booktabs}
\usepackage{adjustbox}
\usepackage{array}
\usepackage{natbib}
\usepackage{multirow}
\usepackage{makecell}
\usepackage{subcaption}
\usepackage{float} 
\usepackage{enumitem}
\title{Can Large Language Models Act as Ensembler for Multi-GNNs?}

\author{
 \textbf{Hanqi Duan\textsuperscript{1}},
 \textbf{Yao Cheng\textsuperscript{1}},
 \textbf{Jianxiang Yu\textsuperscript{1}},
 \textbf{Yao Liu\textsuperscript{1,*}},
 \textbf{Xiang Li\textsuperscript{1,*}}
\\
\\
 \textsuperscript{1}East China Normal University,
\\
 \small{
   \textbf{Correspondence:} \href{mailto:xiangli@dase.ecnu.edu.cn}{liuyao@cc.ecnu.edu.cn} (Yao Liu\textsuperscript{*}) and \href{mailto:xiangli@dase.ecnu.edu.cn}{xiangli@dase.ecnu.edu.cn} (Xiang Li\textsuperscript{*})
 }
}

\newcommand{\ours}{{LensGNN}}
\newcommand{\dhq}[1]{{#1}}

\begin{document}
\maketitle
\begin{abstract}
Graph Neural Networks (GNNs) have emerged as powerful models for learning from graph-structured data.
However,
GNNs lack the inherent semantic understanding capability of rich textual node attributes,
limiting their effectiveness in applications.
On the other hand,
we empirically observe that 
for existing GNN models, no one can 
consistently outperforms others across diverse datasets.
In this paper,
we study whether LLMs can act as an ensembler
for multi-GNNs and propose the \ours~model.
The model first aligns multiple GNNs, mapping the representations of different GNNs into the same space.
Then, through LoRA fine-tuning, it aligns the space between the GNN and the LLM, injecting graph tokens and textual information into LLMs. 
This allows \ours~to ensemble multiple GNNs and take advantage of the strengths of LLM,
leading to a deeper understanding of both textual semantic information and graph structural information.
The experimental results show that \ours~ outperforms existing models.
This research advances text-attributed graph ensemble learning by providing a robust and superior solution for integrating semantic and structural information.
We provide our code and
data here:
\url{https://github.com/AquariusAQ/LensGNN}.
\end{abstract}

\section{INTRODUCTION}

Graphs are structured data that captures the inter-relations between entities in the real world. 
To learn from graphs, 
graph neural networks (GNNs) have been proposed, 
where graph convolution is introduced~\cite{kipf2016semi} and 
message passing is the main mechanism for neighborhood aggregation~\cite{hamilton2017inductive,velivckovic2017graph}.
GNNs have demonstrated significant success across various applications 
such as social network analysis~\cite{hamilton2017inductive}, recommendation systems~\cite{wang2019neural}, and molecular property prediction~\cite{gilmer2017neural}. 

\begin{figure}
    \centering
    \includegraphics[width=0.7\linewidth]{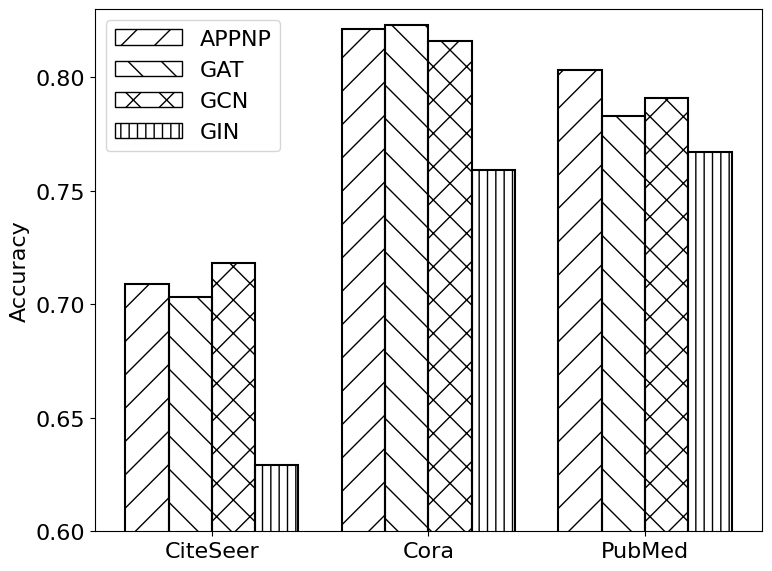}
    % \vspace{-1em}
%     \description{Comparison of node classification accuracy for various GNN models across different datasets. 
% To ensure fairness, 
% the number of hidden layers and the size of hidden representations are kept consistent across these GNNs.}
    \caption{Comparison between GNN models on node classification across different datasets. 
% To ensure fairness, 
% the number of hidden layers and the size of hidden representations are kept consistent across these GNNs. 
}
% \vspace{-1em}
    \label{fig:gnn_accuracies}
\end{figure}

Despite the success, two major challenges remain unresolved. 
First, GNNs, although powerful in capturing graph structures, often lack the inherent semantic understanding capability required to process the rich textual attributes of nodes~\cite{zhao2019semantic}. This can lead to the loss of valuable semantic information during the learning process, limiting the effectiveness of GNNs in applications 
where node features contain meaningful texts, such as citation networks and social media platforms.
Second, 
while there have been a series of GNNs proposed, 
no one has been shown to consistently lead others across datasets of various domains. 
For example, 
we compare the node classification performance of four representative GNNs, namely, 
APPNP~\cite{gasteiger2018predict}, GAT~\cite{velivckovic2017graph}, GCN~\cite{kipf2016semi} and GIN~\cite{xu2019powerfulgraphneuralnetworks}
on three benchmark datasets~\cite{NEURIPS2023_37d00f56}: \emph{Cora}, \emph{Citeseer} and \emph{PubMed}.
For fairness,
we set 
the equal number of hidden representation layers and also the same dimensionality size.
The results given in Figure \ref{fig:gnn_accuracies} demonstrate that 
the winner varies 
across the datasets. 
The challenge of selecting the optimal GNN for a given dataset remains unresolved, 
as different GNN architectures exhibit varying strengths.
This also restricts the wide applicability of GNNs.

To remedy the incapability of GNNs in semantic understanding,  existing works~\cite{qin2024disentangledrepresentationlearninglarge, li2024similaritybasedneighborselectiongraph, wang2024instructgraphboostinglargelanguage, ye2024language, fang2024gaugllmimprovinggraphcontrastive} have resorted to
large language models (LLMs), which
% The recent advent of large language models (LLMs) 
have been shown to be 
prominent in text understanding and generation.
The inherent advantage of GNNs in utilizing graph structure,
fused with the strength of LLMs in semantic understanding, 
mutually enhances both models and leads to superior performance in downstream tasks.
However,
how to select the optimal GNN in different scenarios remains a gap.
Since the effectiveness of GNNs varies across datasets, 
% the study on ensembling GNNs with LLMs is still a gap.
% In this way, 
there naturally arises a question:
\emph{Can we develop an ensembled model for multi-GNNs?}
The ensembled model is expected to integrate the strengths of multiple GNNs and can consistently perform well across datasets.
To further leverage the power of LLMs in text understanding,
we thus upgrade the question: \emph{Can LLMs act as ensembler for multi-GNNs?}

In this paper,
we study whether \textbf{L}LMs can act as an \textbf{ens}embler for multi-\textbf{GNNs} and propose the {\ours} model.
Given multiple GNNs, 
the model ensembles GNNs with LLMs, 
avoiding the tedious efforts in cherry-picking GNN across datasets.
% The model captures semantic information from node textual attributes using LLMs, while simultaneously enabling LLMs to implicitly evaluate and combine the strengths of multiple GNNs.
By dynamically integrating the outputs of different GNNs, 
\ours\ can 
not only preserve the semantic richness of node attributes,
but also optimize the usage 
of diverse GNNs, 
which
enhances the model’s generalizability across a wide range of tasks and datasets.
Specifically,
since multi-GNNs could generate node embeddings in different low-dimensional spaces, we first align multi-GNNs
by alternatively
feeding their generated representations into a shared classifier to
train them sequentially. 
After that,
we freeze the parameters of GNNs to generate node embeddings and
integrate them
into the embedding layer of the used LLM, treating them as embeddings corresponding to graph tokens that encapsulate graph structural information.
Subsequently, we concatenate graph tokens from multi-GNNs, 
text tokens from node attributes, and hand-crafted instructions to create a prompt and fine-tune the LLM using LoRA.
In this way, we can not only align GNNs and LLMs, but implicitly leverage the power of LLM to ensemble multi-GNNs.
Finally,
the main contributions of our paper are given as follows:

\begin{itemize}[leftmargin=*, itemsep=0pt, parsep=0pt, topsep=0pt, partopsep=0pt]
% \begin{itemize}
\item 
We propose a novel method \ours\ to ensemble multi-GNNs with LLMs.
% To our best knowledge,
% we are the first work to attempt the trial.
% \ours\ combines the strengths of both GNNs and LLMs, which
% can be used to avoid the intricate process of manual GNN selection on various datasets. 

\item 
We introduce an effective graph prompt tuning paradigm with both the multi-GNN alignment and the GNN-LLM alignment.
% which enhance LLMs' capability in comprehending graph topology with two phases of alignment:
% the multi-GNN alignment and the GNN-LLM alignment.
% The multi-GNN alignment maps multi-GNNs into the same low-dimensional space while
% the GNN-LLM alignment injects graph token representations into LLMs.

\item We conduct extensive experiments to show the superiority of \ours.
% In particular,
% we compare \ours\ with 17 state-of-the-art competitors. Our experimental results show that \ours\ 
% outperforms
% baselines in a majority of cases, demonstrating its superior performance.

\end{itemize}

\section{RELATED WORK}

GNNs~\cite{kipf2016semi,xu2019powerfulgraphneuralnetworks,velivckovic2017graph,NIPS2017_5dd9db5e} have been widely used in learning from graph-structured data.
% Existing GNN studies have been conducted on various types of graphs such as heterophilic graphs~\cite{zhu2020beyond,pei2020geom,bi2024make}, signed graphs~\cite{derr2018signed,huang2019signed} and temporal graphs~\cite{rossi2020temporal,longa2023graph,xu2020inductive}.
% Further, to improve the effectiveness of GNNs, there are also works~\cite{nguyen2023revisiting,rusch2023survey,black2023understanding} devoted to solving
% the notorious over-smoothing~\cite{chen2020measuring} and over-quashing problems~\cite{akansha2023over}.
% In parallel, to scale GNNs to large-scale graphs, 
% some works~\cite{liao2022scara} focus on enhancing the model efficiency.
% A comprehensive GNN-related survey can be found at~\cite{wu2020comprehensive}.
Despite their advantages, 
GNNs 
struggle to process textual attributes effectively, 
as traditional GNNs lack the semantic understanding capability that is necessary to handle text-attributed graphs (TAGs). 
In addition,
to tailor a GNN model for a given dataset, existing methods mainly rely on neural architecture search (NAS)~\cite{zoph2017neuralarchitecturesearchreinforcement}. However, it is very computationally expensive, which necessitates new exploration. 

Recently, 
LLMs have revolutionized tasks involving semantic understanding and language generation. 
Meanwhile, the ensemble of LLMs and GNNs has garnered significant attention, 
leading to innovative methodologies~\cite{jin2023large}. 
One primary category of approaches takes LLMs as predictors in graph-based tasks, 
% where models 
like GraphGPT~\cite{tang2024graphgptgraphinstructiontuning}, LLaGA~\cite{chen2024llagalargelanguagegraph}
GraphLLM~\cite{chai2023graphllm}
and 
ENGINE~\cite{zhu2024efficienttuninginferencelarge}.
% utilizes few-shot learning to address graph reasoning challenges. ENGINE~\cite{zhu2024efficienttuninginferencelarge} is designed to efficiently integrate textual and topological information by adding a tunable lightweight GNN side structure alongside each layer of LLMs. 
% Similarly, heuristic reasoning approaches, including Chain-of-Thought (CoT)~\cite{wei2022chain} and models like StructGPT~\cite{jiang2023structgpt}, have demonstrated improved graph-based inference capabilities. 
Another notable category is to employ LLMs as encoders, as seen in the works of OFA~\cite{liu2024alltraininggraphmodel}, TextGNN~\cite{zhu2021textgnn} and AdsGNN~\cite{li2021adsgnn}, which focus on optimization strategies for encoding graph data. 
% Employing LLMs as encoders involves using large language models to extract textual features, which are then used as initial feature vectors for nodes or edges in a graph. These feature vectors are passed into graph neural networks (GNNs) for further processing, allowing the model to incorporate both the textual information from the LLM and the structural information from the graph.
Additionally, the alignment of LLMs and GNNs has been explored through prediction alignment methods, such as GOFA~\cite{kong2024gofagenerativeoneforallmodel}, LTRN~\cite{zhang2021minimally} and GLEM~\cite{zhao2022learning}. 
% where iterative training helps improve cross-modality learning. 
% LLMs and GNNs are aligned through iterative pseudo-label training (prediction alignment) or by using contrastive learning to align their latent representations (latent space alignment).
% Finally, graph-empowered LLMs like GreaseLM~\cite{zhang2022greaselm} and DRAGON~\cite{yasunaga2022deep} illustrate the potential of enhancing language models with graph-based knowledge to enrich semantic understanding. They integrate the strengths of large language models and graph structures by modifying Transformer architectures to encode text and graph data simultaneously, allowing for a more comprehensive understanding of the interconnected information within graphs.
These diverse approaches highlight the evolving landscape of LLM and GNN ensemble.
However, 
there still lack studies exploring multi-GNNs ensembling with LLMs,
which is particularly useful
to avoid cherry-picking GNNs for a given dataset.

\section{PRELIMINARIES}

\textbf{Text-attributed graph.} A text-attributed graph is a graph where nodes and edges have textual attributes. In this work, we only consider the case where only nodes have textual attributes. Formally, a text-attributed graph can be defined as \(G = (V, E, X^{(v)})\), where \(V = \{v_1, v_2, \ldots, v_{|V|}\}\) represents the set of nodes, and \(E = \{e_1, e_2, \ldots, e_{|E|}\}\) represents the set of edges, with \(|V| = N\) indicating the total number of nodes in the graph. \(X^{(v)} = \{x^{(v)}_1, x^{(v)}_2, \ldots, x^{(v)}_{|V|}\}\) denote the attributes on nodes, respectively, which are strings. It can be represented as \(G = (V, E, \{x_n\}_{n \in V})\). Additionally, we define \( A \) as the adjacency matrix of graph \( G \), where the matrix size is \( N \times N \). In an unweighted graph, \( A(i,j)=1 \) indicates that there is an edge between nodes \( v_i \) and \( v_j \), while \( A(i,j)=0 \) indicates that there is no edge between them.

\textbf{Graph Neural Networks}. Traditional GNNs are a class of deep learning models designed for handling graph-structured data. The basic architecture of a GNN includes a node representation layer and a message-passing layer. In the node representation layer, each node \(v\) is assigned a feature vector \(x_v\). In the message-passing layer, the representation vector \(h_v^{(t)}\) of node \(v\) is updated after the \(t\)-th iteration using the following formula:

{\small
\begin{equation}
 h_v^{(t)} = \text{UPDATE}\left(\text{AGG}\left(\{h_u^{(t-1)}: u \in \mathcal{N}(v)\}\right), h_v^{(t-1)}\right),
\end{equation}
}

where \(\mathcal{N}(v)\) denotes the set of neighboring nodes of \(v\). The \(\text{AGG}\) function is responsible for aggregating the representations of the neighboring nodes, and the \(\text{UPDATE}\) function combines the aggregated information with the previous state \(h_v^{(t-1)}\) of node \(v\) to update its current state. Through this iterative process, GNNs are able to learn increasingly rich node representations, capturing both local and global structural information in the graph. The specific implementations of the \(\text{AGG}\) and \(\text{UPDATE}\) functions can vary depending on the particular GNN model.

\section{METHODOLOGY}

This section introduces the main steps in \ours,
which includes aligning multi-GNNs and ensembling multi-GNNs with LLM.
The overall model procedure is given in Fig.~\ref{fig:model}.
% To enable the LLM to integrate multiple GNNs, alignment is primarily required in two aspects. 
% First, the output representations of all the GNNs need to be aligned into a common vector space. 
% Second, there must be alignment between the output representations of the GNNs and the semantic space required by the LLM. 
% Based on this strategy, we have designed the model in the following two steps. 
% The structure of the whole model is shown in Fig.~\ref{fig:model}.

\subsection{Aligning multi-GNNs}
Given multiple GNNs, 
they could generate node representations in different low-dimensional spaces.
Therefore, before ensembling GNNs, 
we need to first align them.

\begin{figure*}[ht]
    \centering
    \includegraphics[width=0.9\textwidth]{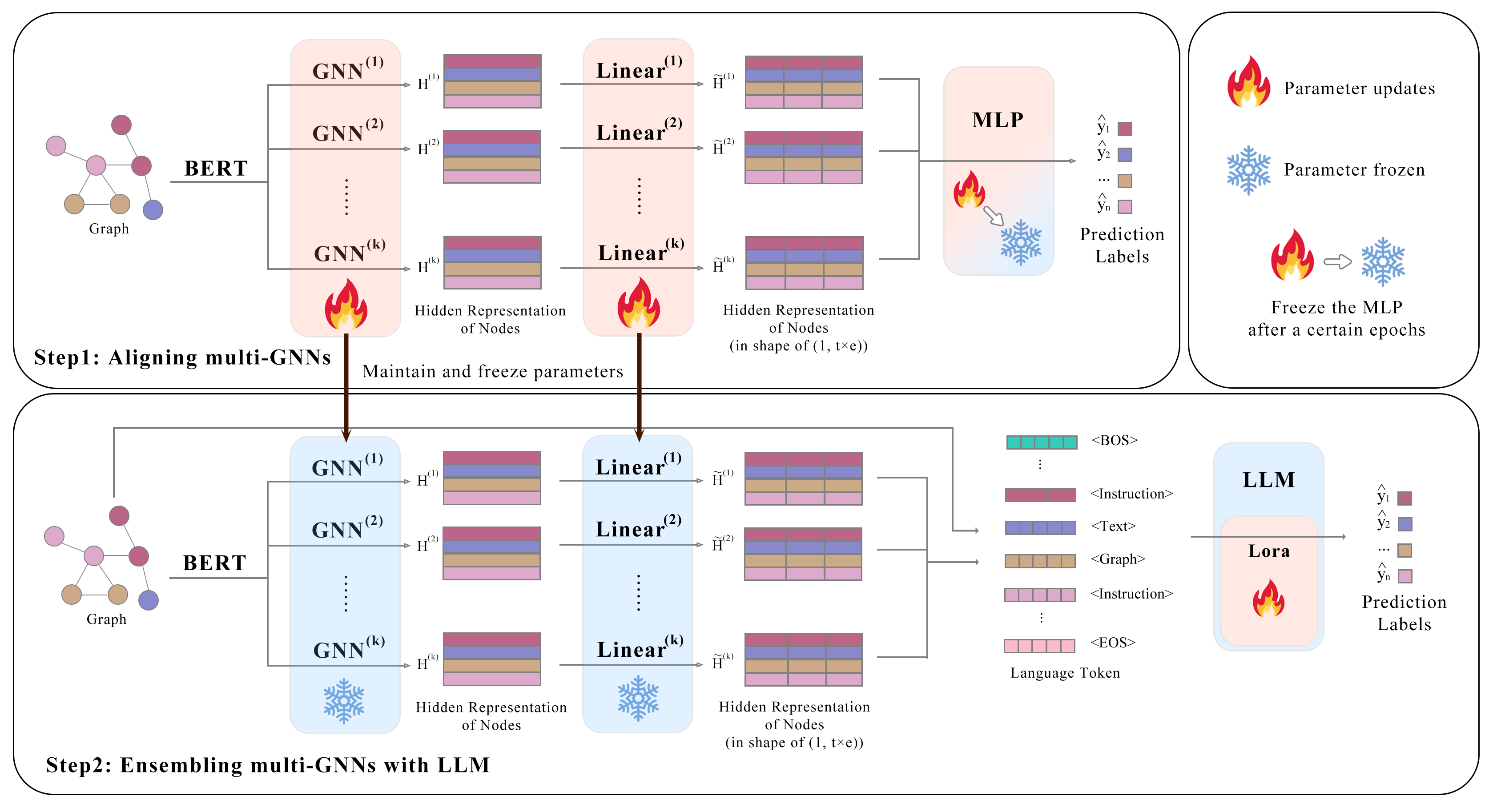}
    % \vspace{-2em}
    \caption{The overall process of \ours.
    % Step 1 is the workflow of the pre-training phase of \ours's GNNs. We employ a language pre-training encoder to extract semantic information from the textual attributes of nodes. Concurrently, we train the GNN encoder to capture structural information. During this phase, a MLP is utilized as a classifier, and the model is trained through gradient feedback. Step 2 is the workflow of the LoRA (Low-Rank Adaptation) supervised fine-tuning stage of \ours's LLM. At this stage, we freeze the parameters of the GNNs and infer the hidden representations of the graph nodes. We segment GNN representations and convert them into graph tokens that LLM can read. The graph tokens, node text tokens, and other prompts are concatenated to form an instruction, which is then input into the LLMs for LoRA supervised fine-tuning.
    }
    % \Description{In the pre-training phase, \ours ~consists of the following architecture: a pre-trained language encoder with frozen parameters, a GNN layer, a linear mapping layer, and an MLP classifier, all of which have parameters that are updated during training.In the fine-tuning phase of \ours, the parameters of the pre-trained language encoder and the GNN layers are frozen, and only the Lora parameters of the LLM are updated. In addition, the size of Hidden Representation is (1, t * e), where e is the length of the enbedding of each token in LLM, and t is the number of tokens each node representation is converted into.}
    \label{fig:model}
\end{figure*}

\textbf{Node feature initialization}. 
To utilize the rich semantic information of textual features,
we use pre-trained language models to initialize them.
Specifically, 
given node $v_i$ with feature vector $x_i$,
we use \texttt{Sentence-BERT}~\cite{reimers2019sentence}
to get its initial embedding vector $\tilde{x}_i$
by:
\begin{equation}
\label{eq:bert}
    \tilde{x}_i = \text{Sentence-BERT}\left(x_i\right). 
\end{equation}
This process allows texts of any dataset and length
to be converted into
the same feature space.

\textbf{GNN representations}. 
Next, the model takes the text representations obtained from Eq.~\ref{eq:bert} for each node and feeds them into multiple GNNs. 
For the $k$-th GNN,
we denote
the node embedding matrix $H^{[k]}$ generated by the final layer
as:
% is given as:
\begin{equation}
    H^{[k]} = \text{GNN}^{[k]}(A, \tilde{X} \mid \Theta_{G_k}),
\end{equation}
where 
$\tilde{X}$ is the initial node embedding matrix from Eq.~\ref{eq:bert} and 
$\Theta_{G_k}$ is the learnable parameters of the $k$-th GNN.
For each node $v_i$,
let $h_i^{[k]}$ be its embedding vector, 
which is the $i$-th row in $H^{[k]}$.

\textbf{Reshape node representations.}
The output of each GNN is then connected to
a linear layer, 
which transforms node representations into the dimensionality of hidden embeddings in LLMs.
This step paves the way for the subsequent alignment of GNN and LLM.
Details will be given in the next section.
For each node $v_i$ in the $k$-th GNN,
its reshaped embedding vector  
$\tilde{h}^{[k]}_i$
is denoted as
\begin{equation}
\tilde{h}^{[k]}_i = \text{Linear}^{[k]}\left(h^{[k]}_i \mid \Theta_{L_k} \right). \\
\label{eq:reshape}
\end{equation}
Note that
the shape of \( \tilde{h}^{[k]}_i \) is a one-dimensional vector of length \( t \times e \), where \( t \) is a hyperparameter indicating the number of graph tokens each node representation is mapped to. 
The dimensionality \( e \) comes from the hidden embedding layer of the LLM used, where each token is mapped to an embedding of shape \( (1, e) \).

\textbf{Multi-GNN alignment}. 
To align node representations from multiple GNNs, 
we next feed them 
into a shared MLP, which serves as the classifier. 
During the training time,
we alternatively input representations from different GNNs into the classifier and train them sequentially. 
This training approach allows node representations from multi-GNNs to be aligned, 
which facilitates the ensemble of multi-GNNs.
For better alignment, we freeze this MLP starting from a certain training epoch.
After training,
the parameters of GNNs and the linear layer in Eq.~\ref{eq:reshape} will be frozen, which are then used
% used to generate node representations 
in aligning GNNs and LLMs.
The overall procedure for GNN alignment is summarized in the top of Fig.~\ref{fig:model}.

In summary, during the multi-GNN alignment step, 
we train each GNN to ensure that their outputs reside in the same vector space. 
By incorporating the node feature initialization step that utilizes language pre-trained models, our model gains the ability to extract semantic and structural information from the nodes in the text graph. 
Building on the full potential of GNNs, we map the dimensions of the GNN representations to the dimensions required by the LLM embeddings, thereby maximizing the ensemble of extensive graph structural information into the LLM in the subsequent steps. 

\subsection{Ensembling multi-GNNs with LLM}
After GNNs are trained, 
we can get 
node representations 
capturing rich graph structural information.
Although
LLMs have 
shown competitive capability of text semantic understanding,
they are ineffective 
in understanding graph structure~\cite{guo2023gpt4graph}.
Therefore,
in the second stage,
we empower LLMs to comprehend graph representations. 
Further, 
with specially designed prompts, 
we enable these models to implicitly perform ensembling. 

\textbf{Align GNNs and LLMs}. 
The key step in enabling LLMs to understand GNN tokens lies in aligning GNN representations, which encapsulate rich structural information, with the semantic space required by LLMs. 
This process necessitates 
fine-tuning LLMs. 
However, due to the extensive number of parameters in these models, fine-tuning requires significant computational resources, 
which is challenging for practical applications.
As a result, 
existing approaches often freeze the parameters of LLMs and train an additional linear layer to act as a projector~\cite{tang2024graphgptgraphinstructiontuning}, mapping GNN representations into the semantic space. 
While the method makes training feasible and yields some effectiveness, 
the expressive capacity of the linear layer is limited, 
leading to the difficulty in
fully leveraging the semantic understanding capabilities of LLMs.

To address the problem,
recent advancements in LoRA (Low-Rank Adaptation)~\cite{hu2021loralowrankadaptationlarge} training have 
been proposed to
directly fine-tune LLMs for alignment.
LoRA is an efficient fine-tuning technique for large pre-trained models,
and introduces low-rank matrices to simulate the effects of full parameter tuning, significantly reducing the number of training parameters and computational resources required. This allows for effective customization of LLMs even with limited resources.
Thus, we adopt LoRA training to fine-tune the LLMs, enabling them to comprehend GNN representations.
Specifically, given an instruction prompt, 
we reserve several \emph{dummy tokens} in the prompt text at the appropriate positions where the GNN representations need to be inserted. 
For example,
in the prompt 
`\textsc{(GNN type: GCN, representations: 
$ <\text{graph\_token}_1>, <\text{graph\_token}_2>, \ldots, <\text{graph\_token}_t> $
), which category does it belong to?}',
\textsc{$<\text{graph\_token}_i>$} denotes the $i$-th dummy token.
Then, for each token in the prompt text,
LLM can generate an embedding vector of size $e$ .
These vectors can form an embedding for the whole prompt text in the format of 
$[\ldots, E'_1, E'_2, \ldots, E'_t, \ldots]$,
where $E'_i$ is the embedding vector of \textsc{$<\text{graph\_token>}_i$}.
After that,
we overwrite the embeddings of dummy tokens by GNN representations.
From Eq.~\ref{eq:reshape},
we can generate a 
GNN representation vector of length \( t \times e \),
which can be further
segmented into \( t \) one-dimensional vectors of length \( e \), denoted as \([E^G_1, E^G_2, \ldots, E^G_t]\).
These vectors serve as the embeddings of \( t \) graph tokens.
After substitution, the raw embedding vector of the prompt text becomes
$[\ldots, E^G_1, E^G_2, \ldots, E^G_t, \ldots]$.
We next feed it into LLM and use LoRA for fine-tuning.
In this way,
we can automatically leverage the language understanding capability of LLM,
and implicitly align structure and semantics.

\textbf{Prompt design and Multi-GNN ensembling}.
After 
aligning graph
structure and semantics,
LLM can understand graph representations.
We next design prompts and 
take LLM as the ensembler to integrate multi-GNNs.
Our goal is to leverage the strengths of both multi-GNNs and LLM to output more accurate predictions.
% The prompt design follows three key principles:

% \begin{enumerate}
%     \item \textbf{Simultaneous input of textual strings from target node and its adjacent neighbors, graph tokens of the node, and task specification.} 
%     Existing studies~\cite{kipf2016semi,velivckovic2017graph,verma2023bet} have demonstrated that aggregating the information from both the center node and its adjacent neighbors could contribute to the label prediction. Further,
%     clear task instructions are necessary for accurate prediction.
%     \item \textbf{ Differentiation between text tokens and graph tokens.} It directs LLM to correctly distinguish between textual tokens and graph tokens within different segments of the prompt, thus preventing confusion.
%     \item \textbf{Guidance for learning from multi-GNNs:} It 
%     leads LLM to implicitly learn how  
%     to effectively combine strengths of different GNNs.
% \end{enumerate}

% When designing prompts, 
% we combine the following components:
The designed prompts include three major components:
% \begin{enumerate}
\emph{task instruction}, 
\emph{node text} and 
\emph{graph token}.
Specifically,
task instruction consists of task specifications and also instructions that guide LLMs to ensemble multi-GNNs.
Node text is used to inject textual attributes of target nodes and their neighbors.
% Additionally, the input incorporates texts from
Due to the input length limitation, for each target node,
we only consider up to 20 adjacent nodes. When the number of adjacent nodes for the target node exceeds 20 or when the total token count of the prompt surpasses the predefined limit, a random sampling of the adjacent nodes is performed.
% , we guide the LLM to distinguish between text tokens and graph tokens, and facilitates the LLM's learning from multiple GNNs, thereby enabling the ensemble of these GNNs.
% \textbf{Node text:} Textual strings from the target node and its adjacent neighbors.
% \textbf{Graph token:} 
Further,
graph tokens are derived from multi-GNNs, which contain rich graph structural information.
After prompts are deigned,
they are fed into LLMs to predict labels.
Formally, we have:
% \end{enumerate}
% we concatenate task instructions, 
% node attributes and graph tokens derived from multiple GNNs.
% Specifically,
% we incorporate textual strings from the target node and its adjacent neighbors, graph tokens of the node, and task specifications. 
% We further guides the LLM to distinguish between text tokens and graph tokens, and facilitates the LLM's learning from multiple GNNs, thereby enabling the integration of these GNNs.
% The ensembling process of LLM is shown in the following formula: 
% we have:
% Based on this, we derive the following formula:
\[
o = LLM(x_{\text{instruction}}, x_{\text{node-text}}, \{x_{\text{GNN}_i}\}_{i=1}^k),
\]
where
\( o \) is the predicted label,
\( x_{\text{instruction}} \) denotes the instruction text, \( x_{\text{node-text}} \) is node textual descriptions and \( x_{\text{GNN}_i} \) denotes the graph representation (graph token) generated by the \( i \)-th GNN. 
% Based on the above principle,
% We provide a prompt template used in our experiments.
% , as shown in Figure~\ref{fig:enter-label}.
% Details on the prompt are given in the caption of the figure.
Details on the prompt template used in our experiments are given in Appendix~\ref{prompt-details}.
Therefore,
% In practice,
the two steps:
{aligning GNNs and LLMs}, and {ensembing  multi-GNNs} can be trained simultaneously.
In our experiments,
we merge them into one step. 
% The prompt can be represented as the concatenation of the following parts:

\section{EXPERIMENT}

% In this section, we conduct extensive experiments to evaluate the performance of our model across various datasets.
% aiming to answer the following questions:

% \begin{itemize}
% \item {\textbf{RQ1}}: How does \ours ~perform compared to SOTA baseline models?
% \item {\textbf{RQ2}}: What is the importance of different components in \ours?
% \item {\textbf{RQ3}}: Can small language models serve as multi-GNN ensembler?
% \item {\textbf{RQ4}}: How efficient is \ours\ ?
% \item {\textbf{RQ5}}: Is our method \ours\ insensitive to model hyper-parameters?
% \end{itemize}

\begin{table*}[h]
    \centering
    
    % \vspace{-0.7em}
    %All figures in the table represent the percentage of accuracy on test dataset.}
    \resizebox{0.8\linewidth}{!}{
    \begin{tabular}{c|c|ccccc}
        \toprule
        \textbf{Model type} & \textbf{Model} & \textbf{Cora} & \textbf{PubMed} & \textbf{ogbn-arXiv} & \textbf{Citeseer} & \textbf{Wiki-CS} \\
        \hline
        MLP & MLP & 66.42 & 82.40 & 61.47 & 71.13 & 68.41\\
        \hline
        \multirow{5}{*}{GNN}
        & GCN & 85.97 & 83.78 & 70.81 & 77.74 & 77.15
 \\
        & GAT & 86.71 & 83.59 & 70.05 & 78.21 & 78.94
 \\
        & GIN & 85.60 & 82.03 & 67.03 & 75.07 & 68.97
 \\
        & GraphSAGE & 84.87 & 87.01 & 70.35 & 73.61 & 79.56\\
        % \hline
        % \multirow{3}{*}{\makecell{GNN \\(Transformer Enhanced)}}
        & Graphormer &  80.41 & 88.75 & 72.81 & 71.28 & 72.07\\
        % & GT & 84.32 & 87.77 & - & 72.51 & 84.05 \\
        % & CoarFormer & 88.69 & 89.75 & 71.66 & \underline{79.20} & - \\
        \hline
        \multirow{5}{*}{Language Model}
        & BERT & 80.15 & 93.91 & 72.78 & 73.17 & 78.33\\
        & SentenceBERT & 78.82 & 92.49 & 71.42 & 72.79 & 77.92\\
        & DeBERTa & 77.79 & 93.45 & 72.90 & 73.13 & 75.11\\
        & SciBERT	& 83.21	& 95.26	& 73.11	& 77.74	& 76.83\\
        & MedBERT	& 77.31	& 93.94	& 72.09	& 74.29	& 74.04\\
        % lm based from Engine
        \hline
        \multirow{4}{*}{
        % \makecell{LLM as Predictor: \\ Graph As Sequence}}
        \makecell{LLM Ensembler}}
        % & DGTL & 81.10 & 87.10 & - \\
        % & SNS-GPT4 & 82.50 & 93.80 & 74.40 \\
        % % & InstructGraph-INS & 89.33 & 81.09 & - \\
        % % GraphText & 87.11 & - & - \\
        % % Graph Agent & \underline{90.65} & \underline{95.40} & - \\
        % % & InstructGLM & \underline{90.77} & 94.62 & 75.70 \\
        % % \hline
        % % \multirow{2}{*}{
        % % \makecell{LLM as Predictor: \\ Graph-Empowered LLM}}
        % % & ENGINE~\cite{zhu2024efficienttuninginferencelarge} & \underline{91.48} & - & 76.02 \\ 
        % % \hline
        % % \multirow{1}{*}{\makecell{LLM as Encoder}}
        % & GAugLLM
        % % ~\cite{fang2024gaugllmimprovinggraphcontrastive}
        % & - & 83.68 & 74.15 \\
        % % & OFA & 74.76 & 78.25 & 77.51 \\
        % % \hline
        % % LLM Direct Inference 
        % & Baichuan2-13B & 13.65 & 36.04 & 4.79 \\
        % % \hline
        % \cline{2-5}
        % \multirow{4}{*}{\makecell{LLM as Predictor \\ and Ensembler}}
        & {\ours-[GCN+GAT]} & 88.56 & 94.11 & 75.78 & 78.05 & 18.41
\\
        & {\ours-[GCN+GIN]} & \textbf{91.88} & 94.47 & \underline{75.89} & 76.64 & \textbf{83.47} \\
        & {\ours-[GAT+GIN]} & 88.19 & \underline{95.43} & 74.31 & \underline{78.99} & \underline{82.98}
\\
        & {\ours-ALL} & \underline{90.40} & \textbf{95.68} &  \textbf{75.91} & \textbf{79.31} & 81.78
\\
        % & \textbf{\ours} & \textbf{91.88} & \textbf{95.43} &  \\
        \bottomrule
    \end{tabular}
    }
    % \vspace{1em} % 增加垂直空间
    \caption{Comparison on classification accuracy (\%) with MLP, GNN models and LM-based methods. \ours\ utilizes Baichuan2-13B-chat as the backbone LLM. We highlight the best score on each dataset in bold and underline the runner-up's. Here, \ours-ALL refers to \ours-[GCN+GAT+GIN]. }
    \hfill
    
    \label{tab:performance}
\end{table*}

\subsection{Experimental settings}

% \subsubsection{Datasets} 
\noindent\textbf{Datasets.}
% We use eight benchmark datasets whose details are given as
% follows. 
We use eight benchmark datasets, comprising five node classification datasets: Cora, PubMed, ogbn-arXiv, Citeseer and Wiki-CS, and three molecular graph classification datasets~\cite{wu2018moleculenet}: BACE, BBBP and ClinTox. 
Details on these datasets are provided in the Appendix~\ref{all-datasets}.

% \noindent\textbf{\emph{Cora}} is a citation network containing 2,708 research papers.
% These papers are categorized into seven classes.
% In this graph, 
% each node represents a scientific paper, and each edge represents a citation. 
% Bag-of-words representation is used as the feature vector for each node, which
% % Each node 
% is also assigned a label indicating its research field. 
% The task for this dataset is node classification.

% \noindent\textbf{\emph{PubMed}} is a citation network dataset containing 19,717 research papers from the biomedical field. 
% These papers are categorized into three classes, and each paper is represented by a bag-of-words feature vector. 
% In this graph, each node represents a scientific paper, and each edge represents a citation. 
% The task associated with this dataset is node classification.

% \noindent\textbf{\emph{ogbn-arXiv}} is part of the Open Graph Benchmark (OGB) and contains 169,343 academic papers scraped from arXiv.
% In this dataset, the nodes represent arXiv papers, and the edges represent citation relationships. 
% These papers are categorized into 40 subject areas. 
% We construct node attributes using the title and abstract of the papers. 
% The task associated with this dataset is node classification.

\noindent\textbf{Baselines.} 
We compare our method with 23 baselines: MLP, 
GNN models, ensemble models, LM-based models and LLM-based models.
Specifically,
GNN models include 
{{GCN}},
% ~\cite{kipf2016semi}, 
{{GAT}},
% ~\cite{velivckovic2017graph},
{{GIN}},
% ~\cite{xu2019powerfulgraphneuralnetworks},
{{GraphSAGE}}~\cite{hamilton2017inductive}
and {{Graphormer}}~\cite{ying2021transformers}.
% For fairness, 
% For these GNN models,
% we use node representations obtained from a pre-trained LM as input.
{Ensemble models include
{{Bagging}}~\cite{breiman1996bagging}, 
{{Stacking}}~\cite{wolpert1992stacked} and
{{AdaBoost}}~\cite{freund1997decision}.}
LM-based models consist of {{BERT}}~\cite{devlin2018bert},
{{SentenceBERT}}~\cite{reimers2019sentence},
{{DeBERTa}}~\cite{he2020deberta}, 
{{SciBERT}}~\cite{beltagy-etal-2019-scibert} and
{{MedBERT}}~\cite{medbert}.
% Further, 
LLM-based models include {{DGTL}}~\cite{qin2024disentangledrepresentationlearninglarge}, {{SNS-GPT4}}~\cite{li2024similaritybasedneighborselectiongraph}, {{GAugLLM}}~\cite{fang2024gaugllmimprovinggraphcontrastive}
and {{Baichuan2-13B}}~\cite{yang2023baichuan2openlargescale}, which all employ 13B LLMs.
% For fairness,
We also compare \ours\ with other 7B-LLM-based models: 
{{GraphGPT}}~\cite{tang2024graphgptgraphinstructiontuning},
{{LLaGA}}~\cite{chen2024llagalargelanguagegraph},
% (17) \textbf{{InstructGraph-INS}}~\cite{wang2024instructgraphboostinglargelanguage}, 
{{InstructGLM}}~\cite{ye2024language},
{{OFA}}~\cite{liu2024alltraininggraphmodel}
and {{GOFA}}~\cite{kong2024gofagenerativeoneforallmodel}.
Due to the space limitation, 
details on these baselines are given in Appendix~\ref{baselines}.

\noindent\textbf{Setup.} 
% In our experiments,
% We implement \ours\ by PyTorch. 
We use Baichuan2
% ~\cite{yang2023baichuan2openlargescale}
and InternLM2.5~\cite{cai2024internlm2} as the backbone LLM for \ours.
In the experiments,
we ensemble
% \ours\ ensembles 
% multiple 
three widely adopted GNN models: GCN, GAT and GIN, each with two layers. 
% Based on the validation set results, we conducted a grid search for the hyperparameters used in model fin-tuning. 
% We perform grid search to fine-tune hyperparameters based on the validation set. 
% Details on the search space is given in Appendix~\ref{app-hyperparameters}. 
% Some fixed LoRA settings include capping each training sample at 2,047 tokens and using half-precision (FP16) for LoRA fine-tuning, with a batch size of 4 per GPU and gradient updates every step. 
% We utilize a cosine-type learning rate scheduler and set the warmup ratio to 0.1.
% Specifically, we utilize LoRA+~\cite{hayou2024lora+}. 
% For the training of LLMs, we utilize Llama Factory~\cite{zheng2024Llamafactory} as a framework.
For GNN models, 
% following [xxx],
we use node representations obtained from the pre-trained {SentenceBERT}~\cite{reimers2019sentence} as input.
% \dhq{The ensemble experiments for more GNNs are presented in Appendix~\ref{app:more-gnn}.}
The additional hyperparameter analysis experiment is detailed in Appendix~\ref{app:add-hyperparameter-analysis}.
The analysis of input length is presented in Appendix~\ref{app-input-length}.
For more details on experiment setup, refer to Appendix~\ref{app:setup}.
Detailed hyperparameter settings are provided in Appendix~\ref{app-hyperparameters}.
Additionally, in Appendix~\ref{app:llama2}, we include supplementary experiments utilizing LLama2-7B as the backbone.
% For baselines that report results on the adopted datasets, we directly report the results from their original papers.
% For those whose results are missing, we leave them blank.
% Due to computational power limitations, for those datasets where the original papers of the baselines did not report results, we left them blank in the table. 
% For Baichuan2, 
% we directly use it without fine-tuning.
% We run all the experiments on a server with a single NVIDIA Tesla A100 GPU. 
% Additionally, due to the high computational cost of training LLM, the results we report are all from single runs.
% \end{comment}

\subsection{Node classification results}

Since the large number of baselines report their results on different datasets and settings,
% Therefore, 
% Due to the large number of baselines,
we show the classification results in three tables, respectively.
Table~\ref{tab:performance} compares \ours\ with MLP, GNN models and small language models.
Tables~\ref{tab:performance-A}
summarizes the results against LLM-based models. Table~\ref{tab:performance-20shot} evaluates the performance of \ours\ in the 20-shot setting.
{Table~\ref{tab:model_ensemble} provides a comparison between \ours\ and methods that ensemble GCN, GAT and GIN.}
% We also experimented with traditional methods that ensemble LLM and GNNs. The experimental analysis is provided in Appendix~\ref{app:traditional-llm-gnn}.
% Table~\ref{tab:performance} shows the classification accuracy of all the methods.
Additionally, we conduct one-shot experiments, which are discussed in Appendix~\ref{app:one-shot}. 
From these tables, we observe:

(1) \ours\ consistently outperforms all the GNN models and LM-based methods across all the datasets.
% This verifies that both semantic 
For GNN models, while they initialize node features with SentenceBERT to capture text semantics,
LLMs have much stronger capability in text understanding and generation,
which explains the advantage of \ours.
For LM-based models,
in addition to the weak expressiveness of small language models,
they ignore the rich structural information of graphs, degrading their performance.

(2) \ours\ achieves better performance than LLM-based models, even in the 20-shot setting.
In particular,
we see that 
Baichuan2-13B performs very poorly. 
We speculate this is because LLMs that have not been fine-tuned are not suitable for node classification tasks on graph data.
Different from other LLM-based models,
\ours\ ensembles the strengths of multiple GNNs
and employs two-phases of alignment, which explains its superior performance. 

{(3) \ours\ leads standard ensemble methods across all datasets. 
% For these ensemble methods,
This is because
when ensembling multiple GNNs, standard ensemble methods can only
leverage graph structural information.
However,
LLMs as ensembler have strong semantic understanding capability that further improves the model performance.

% these methods can only ensemble GNNs that leverage graph structural information but ignore the rich textual semantic information.

% Compared with these methods,
% \ours\ can better capture the rich semantic and structural information, while
% ensemble methods
% ignore the rich semantic information.
% This demonstrates that traditional ensemble techniques, while effective in combining multiple models, are limited in capturing the rich structural and text information of graphs. \ours's superior performance highlights its ability to better integrate and leverage graph-specific patterns compared to standard ensemble approaches.}

\begin{table}
        \centering
        % \vspace{-1em} % 增加垂直空间
        % \vspace{-0.7em}
        \resizebox{1.0\linewidth}{!}{
        \begin{tabular}{c|ccc}
            \toprule
            \textbf{Model} & \textbf{Cora} & \textbf{PubMed} & \textbf{ogbn-arXiv} \\
            \hline
            GraphGPT-MIX-7B & - & 74.16 & 64.76  \\
            LLaGA-HO-7B(GENERAL) & - & 94.45 & 75.01 \\
            InstructGLM-Llama-7B & 87.08 & 93.84 & 75.70 \\
            \hline
            DGTL & 81.10 & 87.10 & - \\
            SNS-GPT4 & 82.50 & 93.80 & 74.40 \\
            GAugLLM & - & 83.68 & 74.15 \\
            Baichuan2-13B & 13.65 & 36.04 & 4.79 \\
            \hline
            \ours*-InternLM2.5-7B & \underline{90.03} & \underline{95.13} & 74.24  \\
            \ours*-Baichuan2-7B & 89.85 & 95.08 & \underline{75.88} \\
            \ours*-Baichuan2-13B & \textbf{91.88} & \textbf{95.68} & \textbf{75.91} \\
            % & \textbf{\ours} & \textbf{91.88} & \textbf{95.43} &  \\
            \bottomrule
        \end{tabular}
        }
        \caption{Comparison on classification accuracy (\%) with LLM-based models.
        % The performance comparison between \ours\ and other open-source LLMs based models. 
        % These models utilize open-source LLMs with 13B parameters or closed-source LLMs as backbone. 
        % These models utilize LLMs as backbone. 
        * denotes the best variant of our model and - means that the results are missing from their original papers.
        {The best variant model is given in Appendix \ref{app:best-variants}.}
        % We highlight the best score on each dataset in bold and underline the runner-up's.
        }
        \label{tab:performance-A}
\end{table}

\begin{table}
        \centering
        % \vspace{0.5em} % 增加垂直空间
        % \vspace{-0.9em} % 增加垂直空间
        % \vspace{-0.7em}
        \resizebox{0.85\linewidth}{!}{
            \begin{tabular}{c|ccc}
            \toprule
            \textbf{Model} & \textbf{Cora} & \textbf{PubMed} \\
            \hline
            GCN & 73.5	& 68.0 \\
            GAT & 72.8 & 68.1 \\
            GIN & 75.7 & 69.3 \\
            \hline
            OFA & 75.34 & 77.89 \\
            GOFA & 77.08 & 87.33 \\
            \hline
            \ours*-Baichuan2-7B & \underline{78.09}& \textbf{89.19} \\
            \ours*-InternLM2.5-7B & \textbf{79.93} & \underline{87.75}  \\
            \bottomrule
            \end{tabular}
            }
        \caption{Classification accuracy (\%) in 20-shot setting.
        % performance(\%) of  LLMs based graph models on the 20-shot  node classification task.
        % We highlight the best score on each dataset in bold and underline the runner-up's.
        }
        \label{tab:performance-20shot}
\end{table}

\begin{table}
\centering
\resizebox{0.95\linewidth}{!}{
\begin{tabular}{c|cccc}
\toprule
\textbf{Model} & \textbf{Cora} & \textbf{PubMed} & \textbf{Citeseer} & \textbf{WikiCS} \\ 
\hline
% GCN            & 0.8721        & 0.8256          & 0.7743            & 0.7630          \\ 
% GAT            & 0.8644        & 0.8115          & 0.7835            & 0.7571          \\ 
% GIN            & 0.8598        & 0.8106          & 0.7093            & 0.7179          \\ 
% \hline
\text{Bagging} & 87.16 & 82.67 & 77.35 & 76.97 \\ 
\text{Stacking} & 86.25 & 84.85 & 78.08 & 74.12 \\ 
\text{AdaBoost} & 84.91 & 83.36 & 76.09 & 75.41 \\ 
\hline
\text{LensGNN-ALL} & \textbf{91.88} & \textbf{95.68} & \textbf{79.31} & \textbf{83.47}     \\ 
\bottomrule
\end{tabular}
}
\caption{Comparison with ensemble methods (\%).}
\label{tab:model_ensemble}
\end{table}

\subsection{Graph classification results}

\begin{table}
        \centering
        % \vspace{-1em} % 增加垂直空间
        % \vspace{-0.7em}
        \resizebox{1\linewidth}{!}{
            \begin{tabular}{c|cc|cc|cc}
            \toprule
            \multirow{2}{*}{\makecell{\textbf{Model}}} & \multicolumn{2}{c|}{\textbf{BACE}} & \multicolumn{2}{c|}{\textbf{BBBP}} & \multicolumn{2}{c}{\textbf{ClinTox}} \\ 
            \cline{2-7}
            & ACC & AUC & ACC & AUC & ACC & AUC \\
            \hline
            GCN & 74.01 & 84.92 & 84.14 & 79.00 & 90.93 & 87.23 \\
            GAT & 74.67 & 80.85 & 76.34 & 66.12 & 91.61 & 86.37 \\
            GIN & 68.09 & \textbf{85.66} & 83.65 & 79.02 & 86.91 & 85.80 \\
            % \hline
            % \ours\-w/o GNN & 77.96 & 77.98 & 86.58 & 81.60 & 98.99 & 95.81 \\
            % \ours-GCN & 76.64 & 76.41 & 86.58 & 81.96 & \textbf{99.32} & 96.00 \\
            % \ours-GAT & 75.98 & 75.07 & 86.34 & 80.73 & 98.65 & 95.63 \\
            % \ours-GIN & 69.73 & 68.80 & 87.80 & 84.54 & 98.32 & 91.81 \\
            \hline
            \ours-[GCN+GAT] & 73.02 & 72.67 & 86.09 & 81.29 & 98.99 & 94.00 \\
            \ours-[GCN+GIN] & 75.32 & 74.59 & 88.04 & 83.27 & \textbf{99.32} & 96.00 \\
            \ours-[GAT+GIN] & 76.31 & 75.99 & 86.09 & 81.64 & 98.99 & 95.81 \\
            \ours-ALL & \textbf{80.59} & 80.33 & \textbf{88.53} & \textbf{85.37} & 98.99 & \textbf{97.63} \\
            \bottomrule
            \end{tabular}
            }
        \caption{Graph classification results.}
        \label{tab:performance-E}
\end{table}

We further evaluate the effectiveness of \ours\ on the graph classification task.
We use classification accuracy (ACC) and AUC scores as the evaluation metrics.
The results are given in Table ~\ref{tab:performance-E}. From the table,
% we see that
\ours\ outperforms GNN baselines in most cases, which shows that combining GNNs and LLMs is inspiring.
Further, 
\ours-ALL generally performs better than its variants.
This verifies that leveraging strengths of multiple GNNs is useful.
Surprisingly, we find that on BACE, 
\ours\ and its variants achieve smaller AUC scores than GNN models.
We step into the datasets and observe that 
BACE is a label-balanced dataset while BBBP and ClinTox are highly imbalanced.
Note that, for balanced dataset, ACC is more convincing than AUC score, while AUC score is a better indicator than ACC on imbalanced dataset.
Therefore,
the ACC advantage of \ours\ on BACE and the AUC leads on other two datasets
have evidently demonstrated the superiority of our method.

% To validate that our model not only performs well in node classification, we conducted graph classification tasks and present the results in Table ~\ref{tab:performance-E}. 
% The three categories of models in the table are GNN models, \ours\ aligned with only a single GNN, and \ours\ that integrates multi-GNN models.
% From the table, we observe that:

% (1) On molecular graph classification tasks, \ours\ that ensembles multiple GNN models achieves the best classification accuracy. For GNN models, similar to the situation in node classification, they fail to efficiently utilize semantic information. For \ours\ aligned with only a single GNN, due to the lack of integration of multiple GNNs, they obtain limited structural information. This demonstrates that \ours\ relies on the integration of multiple GNNs to acquire powerful learning capabilities.

% (2) For imbalanced datasets, \ours\ also demonstrates excellent performance. In experiments with label-imbalanced datasets BBBP and ClinTox, measured by ROC-AUC, \ours\ also achieved the most outstanding results. We speculate that this is attributed to the strong generalization capabilities of LLMs, which enhance \ours's ability to comprehend minority classes when integrating multi-GNNs.

\subsection{Ablation study}

% In this section, 
We next systematically conduct an ablation study on Cora, PubMed, and ogbn-arXiv datasets to evaluate the importance of major model components.
Specifically,
we explore various configurations of {``GNN Encoder''}~(using different GNNs
), 
{``Alignment''}~(whether multiple GNNs are aligned when training GNNs
), 
{``With Text''}~(whether node text is included in prompt)
and {``With Neighbor''}~(% the number of neighbor hops of the node text when inputting LLMs
whether neighborhood information is included or not).
The results are presented in Table~\ref{tab:ablation}.
% From the table, we see that:

\textbf{Performance of different GNN encoders.}
Ensembling multiple GNNs noticeably outperforms using a single GNN. 
For example, on the ogbn-arxiv dataset, the best result from a single GNN is 74.35\%, 
while \ours~achieves 75.91\%. 
This illustrates that by integrating the strengths of various GNNs, 
\ours~can effectively enhance the overall performance.

On Cora and Wiki-CS, the small training sets introduce more noise during training. Since our model relies on an LLM as backbone—and LLMs typically require large-scale data for effective learning—we attribute the underperformance of LensGNN-ALL compared to LensGNN-[GCN+GIN] primarily to the limited training data and elevated noise levels. These factors hinder the LLM from adequately learning to align all three GNNs (GCN, GAT, and GIN) simultaneously.

\textbf{Impact of Alignment.}
On the PubMed dataset, the results for unaligned GNN and aligned GNN are 93.96\% and 95.68\%, respectively, while on the ogbn-arXiv dataset, the results are 73.73\% and 75.91\%, respectively.
This shows that multi-GNN alignment can help LLMs better understand 
graph tokens.
Although the performance of aligned GNN is slightly lower than unaligned GNN on the Cora dataset, 
this may be due to the insufficient number of training samples.

\textbf{Importance of Node Text.}
It can be observed that node text plays a key role in the supervised fine-tuning of LLMs.
Through these node texts,
the model can achieve a deeper understanding of semantics, resulting in outstanding accuracy.

% \emph{\textbf{Role of Neighbors.}} 
\textbf{Role of Neighbors.}
The textual information of neighbors provides support for LLM’s understanding of the semantic information of nodes, 
% enabling LLM to grasp structural information more deeply, 
thereby improving classification performance.

% Overall, the ablation study confirms that each component of the model contributes significantly to its overall performance. 
\begin{table*}[h]
    \centering
    \small
    % \Description{
    % These elements enhance the model's ability to integrate representations and contextual information, resulting in higher accuracy across various datasets.
    % }
    % \vspace{-1em}
    \resizebox{0.95\linewidth}{!}{
    \begin{tabular}{lccc|ccc}
        \toprule
        \textbf{GNN Encoder} & \textbf{Alignment} & \textbf{With Text} & \textbf{With Neighbor} & \textbf{Cora (Acc/\%)} & \textbf{PubMed (Acc/\%)}  & \textbf{ogbn-arXiv (Acc/\%)}  \\
        \hline
        -   & -       & Yes & 1 & 87.82 & 93.77 & 74.47 \\
        GCN & -       & Yes & 1 & 89.29 & 94.23 & 74.35 \\
GAT & -       & Yes & 1 & 90.03 & 94.26 & 73.37 \\
        GIN & -       & Yes & 1 & 88.92 & 94.82 & 73.26 \\
        \hline
        GCN, GAT, GIN & Yes & No & 1 & 19.92 & 40.06 & 34.92 \\
        \hline
        GCN, GAT, GIN & No & Yes & 1 & \textbf{90.77} & 93.96 & 73.73 \\
        \hline
        GCN, GAT, GIN & Yes & Yes & 0 & 84.87 & 94.21 & 72.03 \\
        \hline
        % GAT, GIN & Yes & Yes & 1 & 8819 & 9543 & 7431 \\
        % GCN, GIN & Yes & Yes & 1 & \textbf{9188} & 9447 & 7589 \\
        % GCN, GAT & Yes & Yes & 1 & 8856 & 9411 & 7578 \\
        % \hline
        GCN, GAT, GIN & Yes & Yes & 1 & 90.40 & \textbf{95.68} & \textbf{75.91} \\
        \bottomrule
    \end{tabular}
    }
    \caption{Ablation study results.}
    \label{tab:ablation}
\end{table*}

\subsection{Performance with different backbones}
\begin{table*}
    \centering
    \small
     % \Description{ The figure indicates that the model possesses the capability to integrate GNN representations when any LLM serves as the backbone of \ours~, whereas this ability is absent when employing language models with fewer parameters.}
    % \vspace{-1em}
    \resizebox{0.95\linewidth}{!}{
    \begin{tabular}{lc|ccc|cccc}
        \toprule
        \textbf{Model} & & \textbf{Bert-base} & \textbf{\makecell{T5-base\\(Encoder only)}} & \textbf{T5-base} & \textbf{Falcon-7B} & \textbf{InternLM2.5-7B-chat} & \textbf{Baichuan2-13B} \\
        \hline
        \textbf{Parameters} &  & 110M & 110M & 220M & 7B & 7B & 13B \\
        \hline
        \multirow{4}{*}{\textbf{Cora}}
        & With GCN & 81.48 & 78.52 & 76.10 & 85.23 & 89.29 & 89.29 \\
        & With GAT & 81.85 & 79.26 & 77.94 & 80.44 & 89.66 & 90.03 \\
        & With GIN & \textbf{87.50} & \textbf{88.24} & \textbf{88.24} & 87.45 & 89.29 & 88.92 \\
        & Ensemble All & 80.37 & 80.00 & 77.57 & \textbf{89.66} & \textbf{90.03} & \textbf{90.40} \\
        \hline
        \multirow{4}{*}{\textbf{Pubmed}}
        & With GCN & \textbf{94.67} & \textbf{94.58} & 94.03 & 94.67 & 94.72 & 94.23 \\
        & With GAT & 94.37 & 94.53 & \textbf{94.19} & 94.37 & \textbf{95.13} & 94.26 \\
        & With GIN & 94.01 & 94.17 & 94.00 & 94.52 & 94.87 & 94.82 \\
        & Ensemble All & 94.17 & 93.51 & 93.95 & \textbf{94.97} & \textbf{95.13} & \textbf{95.68} \\
        \bottomrule
\end{tabular}
}
    \caption{The classification accuracy (\%) with different LLMs on Cora and PubMed. 
    % We highlight the best score on each dataset in bold.
    }
    \label{tab:llms}
\end{table*}
To study the impact of backbone models on \ours, 
we use different backbones, 
including three small LMs: BERT-base~\cite{devlin2018bert}, T5-base~\cite{raffel2023exploringlimitstransferlearning}, the encoder-only variant of T5-base, 
and two other LLMs: Falcon-7B~\cite{falcon40b} and InternLM2.5-7B-chat~\cite{cai2024internlm2}. 
% The experiments based on these LLMs include aligning each language model separately with the representation of each type of GNN, as well as aligning them simultaneously with all three types of GNNs (GCN, GAT, GIN) and performing an ensemble. 
All the results are presented in Table~\ref{tab:llms}.
From the table, we see that

(1) Small LMs BERT-base, T5-base, and T5-base (Encoder only) perform well when applied to a single GNN, but the performance drops significantly when ensembling multiple GNN models. 
This suggests that models with less parameters lack the capability to ensemble multiple GNNs.

(2) 
For LLMs, they generally perform better than LM backbones.
For example, for GAT on Cora, the best result for LM backbones is 81.85\%,
while that for LLM backbones is 90.03\%.
This is because LLMs can provide richer text semantics.
Further,
compared with single GNN,
the ensemble of multiple GNNs with LLMs leads to better performance,
which verifies that unifying the strengths of multiple GNNs is beneficial for node classification.

% In contrast, LLMs with more parameters that ensemble a single GNN generally perform better than smaller LMs in most cases.

% (3) Across different large language models, the ensemble of multiple GNN models achieved the best performance on all datasets.

% Overall, all the observations indicate that LLMs can better achieve the ensemble of GNN models, 
% and the multi-GNNs ensemble can harnesses the strengths of multiple GNNs to achieve the best predictions.

\subsection{Traditional methods ensembling LLM and GNNs }
% \label{app:traditional-llm-gnn}
We conduct experiments employing traditional ensemble learning methods to ensemble the predictions from both the LLM and GNNs (GCN, GAT and GIN) on 4 datasets. The results are as Table~\ref{tab:tra-ensemble-llm-gnn}:

\begin{table}[!hbpt]
        \centering
        \resizebox{0.99\linewidth}{!}{
        \begin{tabular}{l|ccccc}
        \toprule
        \textbf{Method} & \textbf{LLM} & \textbf{Cora} & \textbf{PubMed} & \textbf{Citeseer} & \textbf{Wiki-CS} \\
        \hline

Bagging	& Baichuan2-13B	& 86.90	& 86.16	& 75.55	& 76.47 \\
Stacking	& Baichuan2-13B	& 85.24	& 85.70	& 70.53	& 67.35 \\
        \hline
LensGNN*	& Baichuan2-13B	& \textbf{91.88}	& \textbf{95.68}	& \textbf{79.31}	& \textbf{83.47} \\

        \bottomrule
    \end{tabular}
        }
        \caption{Comparison with traditional ensemble of LLM and GNNs. (\%)}
        \label{tab:tra-ensemble-llm-gnn}
\end{table}

As demonstrated in our experiments, LensGNN consistently outperforms all conventional ensemble methods. 

In addition, the observed phenomenon where ensemble learning results underperform a single LLM (as shown in Table~\ref{tab:ablation}) can be attributed to the fact that traditional ensemble methods may sometimes yield worse performance than individual models. 
% Key reasons include: 1. Strong base learners: Bagging’s random sampling can disrupt the high performance of an already powerful LLM. 2. Large performance gaps: In Stacking, significantly weaker base learners (e.g., GNNs) introduce noise and impair the meta-learner.
Potential reasons for this phenomenon include: 1) Strong base learners: Since the base learner (LLM) itself is already a strong model, the random sampling in Bagging may disrupt its original high performance. 2) Large performance gap among base learners: When base learners differ significantly in performance (e.g., LLM vs. GNN), weaker base learners (GNNs) may introduce noise, negatively affecting the meta-learner in Stacking.

\subsection{Ensemble more GNN}
% \label{app:more-gnn}
We further incorporate a larger number of GNNs to evaluate how \ours\ scales and whether the performance continues to improve. The results are presented in Table \ref{tab:more-gnn}.
Overall, we see that with the increase in the number of GNNs, the model performance improves further. However, from the results on Cora and Pubmed, simply adding GNNs will not bring sustained improvement.
We speculate that this is because more GNNs will introduce more graph tokens and also noise.
The increased input length will
bring difficulty in model learning, so does noise.

\dhq{Based on the empirical findings presented in Table \ref{tab:more-gnn}, we propose a heuristic recommendation for researchers to select 3-4 high-performing GNNs (e.g., GCN, GAT, GIN) when constructing ensembles for new datasets. Notably, the inclusion of additional GNN models incurs minimal computational overhead, allowing for flexible expansion until validation accuracy approaches saturation—the point at which further performance improvements become negligible. This strategy balances model diversity and efficiency while maximizing predictive accuracy.}

\begin{table*}[!hbpt]
        \centering
        \resizebox{0.95\linewidth}{!}{
        \begin{tabular}{l|ccccc}
        \toprule
        \textbf{Variants of \ours} & \textbf{Number of GNNs} & \textbf{Cora} & \textbf{PubMed} & \textbf{Citeseer} & \textbf{Wiki-CS} \\
        \hline
LensGNN-GCN	& 1	& 0.9059	& 0.9490	& 0.7836	& 0.8294 \\
LensGNN-[GCN+GAT]	& 2	& 0.9003	& 0.9503	& 0.7868	& 0.8306 \\
LensGNN-[GCN+GAT+GIN]	& 3	& 0.9114	& 0.9508	& 0.7884	& 0.8334 \\
LensGNN-[GCN+GAT+GIN+GraphSAGE]	& 4	& 0.9095	& 0.9513	& 0.7852	& 0.8332 \\
LensGNN-[GCN+GAT+GIN+APPNP]	& 4	& 0.9040	& 0.9500	& 0.7915	& 0.8366 \\
LensGNN-[GCN+GAT+GIN+GraphSAGE+APPNP]	& 5	& 0.9022	& 0.9472	& 0.7931	& 0.8417 \\

        \bottomrule
    \end{tabular}
        }
        \caption{Classification accuracy when integrating more GNNs. The backbone LLM is Internlm2.5-7B-chat.}
        \label{tab:more-gnn}
\end{table*}
}

\subsection{Model efficiency study}

% We next analyze the computational efficiency of \ours\ during both the training and inference phases.

\textbf{Training Efficiency.} 
The training of \ours\ includes two main phases: aligning multi-GNNs and ensembling multi-GNNs with LLM.
In the first phase,
the time complexity depends on the GNN encoders.
Since the used GNNs have a linear time complexity w.r.t. the number of nodes in the graph, the training process is efficient.
% the training efficiency primarily depends on the efficiency of the GNN encoders utilized, while also being related to the hyperparameter Graph token \( t \).
% Overall, the computational efficiency in this phase is comparable to that of pure GNN training. 
In the second phase, 
fine-tuning LLMs is the main computational cost.
% the training efficiency predominantly relies on the fine-tuning efficiency of the LLMs. 
We employ LoRA to reduce the number of parameters to be fine-tuned. 
For example, 
for Baichuan2-13B, the trainable parameters constitute only 0.0470\% of the total parameters. 
This allows us to fine-tune the 13B parameter LLM in a single 80G Nvidia A100 environment.
% with a training batch size of 4 per device.
% To summarize, 
Compared to GraphGPT~\cite{tang2024graphgptgraphinstructiontuning}, the increase of our training computational cost primarily stems from employing more than one GNN. 
However, in contrast to LLM, the number of parameters in GNN is considerably smaller,
which increases marginal training cost.
% ensuring that the addition of GNN does not significantly impact the computational cost.

\begin{figure}
    \centering
    \includegraphics[width=0.75\linewidth]{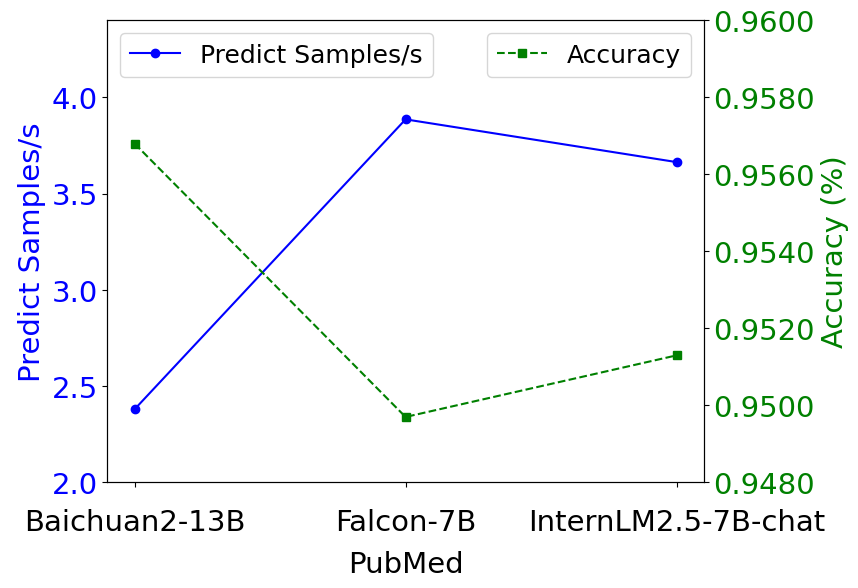}
    % \vspace{-1em}
    \caption{Inference speed and accuracy comparison. 
    % based on different backbone LLMs.
    }
    \label{fig:efficiency}
\end{figure}

\textbf{Inference Efficiency.} 
During inference, 
all the parameters are frozen. 
% In the first phase, the parameter count is relatively low, leading to minimal inference time compared to the second phase. 
The major cost of model inference comes from 
LLMs.
% Since the model parameters in the first stage are much smaller than the LLM in the second stage, the time cost in the inference phase mainly comes from the second stage’s LLM inference.
% In contrast, the inference speed in the second phase is contingent upon the total parameter count of the LLMs. 
In our experiments, with a maximum input length of 2047 tokens, the inference speed and accuracy of our method based on Baichuan2-13B, Falcon-7B, and InternLM2.5-7B-chat on the Pubmed dataset is illustrated in Figure~\ref{fig:efficiency}. 
Overall, the inference speed ranges from $2$ to $4$ samples per second. 
Although a larger LLM results in reduced inference speed, 
it concurrently yields superior accuracy. 
% While LLM introduces additional overhead, 
% it delivers significant performance gains. 
% In accuracy-sensitive domains 
% % like healthcare or finance
% , this trade-off holds practical value.
% In practical deployments, the backbone LLM can be selected by considering the balance between computational power and application requirements.

\subsection{Hyperparameter sensitivity analysis}

\begin{figure}
    \centering
    \includegraphics[width=0.75\linewidth]{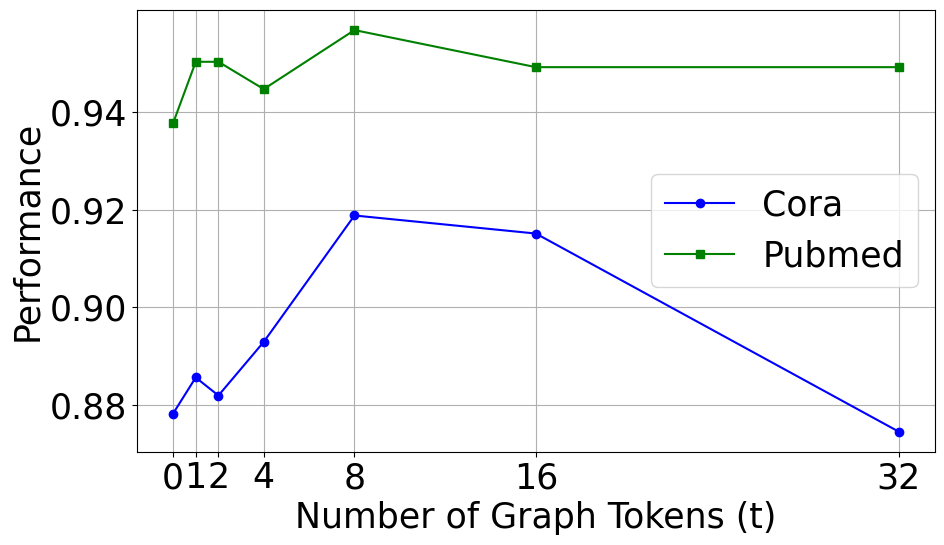}
    % \vspace{-1em}
    \caption{Hyperparameter sensitivity analysis.}
    % (Number of Graph Tokens $t$). 
    % \Description{The figure demonstrates that the performance initially increases and then gradually decreases as $t$ grows, with the optimal performance achieved at $t = 8$.}
    % }
    \label{fig:hp}
\end{figure}

We end this section with model sensitivity analysis on
the number of graph tokens $t$,
% is an important hyperparameter, representing 
which represents how many tokens each node is mapped into before fed into LLM.
We conduct experiments on Cora and Pubmed with varying $t$ 
values,
% set to values of 1, 2, 4, 8, 16 and 32, 
and the results are shown in Figure~\ref{fig:hp}.
From the figure,
% the results indicate that 
we see that, as the number of graph tokens increases, 
the model's performance first rises  and then decreases.
For both datasets,
the best results are achieved when $t=8$.
When $t$ is small, 
graph token representations cannot well capture the semantic information.
% the distinction between nodes becomes insufficient, making it difficult for LLMs to fully learn the semantic information. 
On the other hand, when $t$ is large, the representation of graph tokens could be noise-corrupted, 
which adversely affects the model's understanding on graph structure. 
Therefore, an appropriate number of tokens helps LLMs effectively capture structural information, enhancing semantic comprehension and leading to better performance.
% In addition, we have also supplemented the sensitivity analysis on Citeseer and Wiki CS in the appendix \ref{app:add-hyperparameter-analysis}.

\section{CONCLUSION}

We studied the problem of how to ensemble multi-GNNs in this paper and proposed \ours, which ensembles multi-GNNs with LLMs.
\ours\ adopts two phases of alignment:
the multi-GNN alignment aims to map node representations from different GNNs into the same low-dimensional space, while the 
GNN-LLM alignment injects graph representation as graph tokens into LLM.
After that, we performed  supervised fine-tuning with LoRA to enhance the LLM's capability in understanding graph topology.
Finally,
we conducted extensive experiments to show the effectiveness of our ensembling model \ours.
In particular, 
% our experimental 
the results show that LLMs can serve as effective multi-GNN ensembler, while small LMs cannot.

\clearpage

\section*{Limitations}

% We have several limitations in concern to our method.
Our method could suffer from several limitations:

First, due to limited computational resources, we do not conduct experiments on LLMs with parameters exceeding 13B. This implies that the upper limit of our method's capabilities has not been fully investigated. For larger-scale LLMs that are utilized in production environments, 
% future research can explore the performance of our method.
they could further improve our model's performance.

Second, our method relies on manually crafted prompts to ensemble multi-GNNs. Future studies on prompt design could be a promising direction.
% for specific tasks, in order to ensemble multiple GNNs, it is required to manually design specific prompts. 
% Consequently, the effectiveness may be influenced by human intervention. Researching improvements to such prompts or exploring end-to-end integration methods represents a promising direction.

Third, we only evaluated the performance based on node classification and graph classification tasks.
Evaluation on other tasks, such as graph dataset generation and graph task interpretability analysis, are valuable research questions.
% without fully utilizing the language generation capabilities of LLMs. Whether it is possible to exploit language generation capabilities in the context of integrating LLMs with GNNs to perform a wider range of tasks, such as graph dataset generation and graph task interpretability analysis, constitutes a valuable research question.

% \textbf{Human-designed Prompts.} Our approach provides a framework for harnessing the potential of LLMs to integrate GNNs. However, for specific downstream tasks, the design of prompts to leverage this capability still require human intervention. Further exploration into the existence of end-to-end ensembling methods is a promising direction.

% \textbf{Interpretability.}  The language generation capability of LLMs makes it possible for interpretability in graph tasks, so whether the interpretability of the integration of multi-GNNs can be improved is a valuable research question.

\section*{Ethical Statement}

% After our rigorous review, we are confident that there are no ethical or moral issues involved in this work.
Our work falls under basic research and is not tied to specific applications; therefore, whether our method will be abused and cause negative social impacts depends on the specific applications in which others use our method. In addition, our work does not involve any stakeholders benefiting or being disadvantaged, nor does it involve vulnerable groups. 
The datasets we used are all commonly used public datasets, posing no privacy risks, and aligned with their intention for scientific research. For the datasets, pre-trained models, and training frameworks utilized, we adhered to their respective open-source licenses: community license for Baichuan-2~\cite{yang2023baichuan2openlargescale}, Apache License 2.0 for InternLM2.5~\cite{cai2024internlm2} , Falcon-7B~\cite{falcon40b} and Llama-Factory~\cite{zheng2024llamafactory}

\section*{Acknowledgments}

This work is supported by National Natural Science Foundation of China No. 62202172 and No. 42375146.


% \clearpage
\bibliography{ref}

\clearpage
\appendix

\section{Datasets}
\label{all-datasets}
% \begin{table*}[h]
% \centering
% \caption{Datasets used in the paper. Summary of datasets used in the study, including their year of release, task type, number of nodes and edges, and domain.}
% \label{tab:datasets}
% \begin{tabular}{lccccc}
% \toprule
% \textbf{Dataset} & \textbf{Year} & \textbf{Task} & \textbf{\# Nodes} & \textbf{\# Edges} & \textbf{Domain} \\
% \midrule
% Cora & 2000 & NC & 2,708 & 5,429 & Academic \\
% PubMed & 2020 & NC & 19,717 & 44,338 & Academic \\
% ogbn-arXiv & 2020.5 & NC & 169,343 & 1,166,243 & Academic \\
% \bottomrule
% \end{tabular}
% \end{table*}
We use eight benchmark datasets, comprising five node classification datasets: Cora, PubMed, ogbn-arXiv, Citeseer and Wiki-CS, and three molecular graph classification datasets: BACE, BBBP and ClinTox. 
Statistics of these datasets are summarized in Table~\ref{tab:datasets}.

% \begin{table}[h]
% \centering
% \caption{
% Datasets statistics.
% }
% % \vspace{-0.5em}
% \label{tab:datasets}

% \begin{subtable}[t]{\linewidth}
% \centering

% \caption{Datasets for node classification.}
% \vspace{-0.5em}
% \label{tab:datasets-a}

% \begin{tabular}{lccc}
% \toprule
% \textbf{Dataset} & \textbf{\#~Nodes} & \textbf{\#~Edges} & \textbf{\#~Classes}  \\
% \midrule
% Cora & 2,708 & 5,429 & 7  \\
% PubMed & 19,717 & 44,338 & 3  \\
% ogbn-arXiv & 169,343 & 1,166,243 & 40 \\
% Citeseer & 3,312 & 4,732 & 6 \\
% Wiki-CS & 11,701 & 216,123 & 10 \\
% \bottomrule

% \end{tabular}
% \end{subtable}
% \begin{subtable}[t]{\linewidth}
% \centering

% \vspace{0.5em}

% \caption{Datasets for graph classification.}
% \vspace{-0.5em}
% \label{tab:datasets-b}

% \begin{tabular}{lccc}
% \toprule
% \textbf{Dataset} & \textbf{\#~Molecules} \\
% \midrule
% BACE &  1,513  \\
% BBBP & 2,039   \\
% % Tox21 & 7,831 \\
% ClinTox & 1,491 \\
% % SIDER &  1,427 \\
% \bottomrule

% \end{tabular}
% \end{subtable}

% \end{table}

\noindent\textbf{\emph{Cora}}~\cite{mccallum2000automating} is a standard citation network dataset consisting of 2,708 research papers in the field of machine learning. 
% These papers are categorized into seven classes, and each paper is represented by a 1,433-dimensional bag-of-words feature vector, reflecting the presence of words in the titles and abstracts. 
These papers are categorized into seven classes. 
The title and abstract of each paper are utilized as the textual attributes of the nodes.
The dataset includes 5,429 citation links, constructing a graph structure among the papers. The Cora dataset is commonly used for node classification and evaluating the performance of graph neural network models.

\noindent\textbf{\emph{Citeseer}}~\cite{giles1998citeseer} is a citation network dataset comprising 3,312 research papers primarily from the fields of computer science and information technology. These papers are categorized into six classes based on their research areas. The title and abstract of each paper are used as the textual attributes of the nodes, providing a semantic representation of the content. The dataset includes 4,732 citation links, forming a graph structure that represents the citation relationships between the papers. The Citeseer dataset is widely used for node classification tasks, and it serves as a benchmark for evaluating the performance of graph neural network models.

\noindent\textbf{\emph{PubMed}}~\cite{sen2008collective} is another citation network dataset containing 19,717 research papers from the biomedical field. These papers are categorized into three classes.
% and each paper is represented by a 500-dimensional bag-of-words feature vector, based on the words in the abstracts. 
The title and abstract of each paper are utilized as the textual attributes of the nodes.
The dataset includes 44,338 citation links, constructing a graph structure among the papers. The PubMed dataset is also used for node classification tasks, particularly in testing large-scale graph neural network models.

\noindent\textbf{\emph{ogbn-arXiv}}~\cite{hu2020open} is part of the Open Graph Benchmark (OGB) and contains 169,343 academic papers scraped from arXiv. These papers are time-ordered by their submission dates and categorized into 40 subject areas. In this dataset, the nodes represent arXiv papers, and the edges represent citation relationships. The dataset includes 1,166,243 citation links, constructing a citation network among the papers. The ogbn-arXiv dataset is commonly used for node classification and studying time-sensitive graph neural network models.

\noindent\textbf{\emph{Wiki-CS}}~\cite{mernyei2022wikicswikipediabasedbenchmarkgraph} is derived from Wikipedia and consists of 11,701 web pages related to computer science topics. These pages are categorized into 10 classes, each representing a different area of computer science such as artificial intelligence, computer architecture, and software engineering. The text content of each page is used as the textual attributes of the nodes, capturing the thematic essence of the pages. The dataset includes 216,123 hyperlinks, which construct a graph structure connecting the web pages. The Wiki-CS dataset is often utilized for node classification tasks and for evaluating graph-based machine learning models.

\noindent\textbf{\emph{BACE}}~\cite{wu2018moleculenet} is a collection of inhibitors of human beta-secretase 1 (BACE-1) and provides both quantitative IC50 values and qualitative binary labels indicating the binding results. The BACE dataset comprises 1,513 molecular graphs for the molecular property prediction.

\noindent\textbf{\emph{BBBP}}~\cite{wu2018moleculenet} is used for predicting blood-brain barrier permeability (BBBP), which is crucial for determining whether a molecule can cross the blood-brain barrier. The BBBP dataset contains 2,039 molecular graphs for the binary graph classification task.

\noindent\textbf{\emph{ClinTox}}~\cite{wu2018moleculenet} is a collection of drugs approved by the FDA and those that have failed clinical trials due to toxicity reasons. The dataset encompasses two classification tasks for 1,491 drug molecules: (1) clinical trial toxicity and (2) FDA approval status.

For Cora, PubMed, Citeseer, BACE, BBBP and ClinTox, we randomly split nodes into 60\%,
20\%, and 20\% for training, validation and testing, and measure the performance of all models on the test set. 

For ogbn-arXiv we split the dataset as suggested in~\cite{hu2020open}.

For Wiki-CS, we split the dataset as suggested in~\cite{mernyei2022wikicswikipediabasedbenchmarkgraph}.

\begin{table}[]
\scriptsize
\centering
% \vspace{-0.5em}
\begin{tabular}{@{}ccccc@{}}
\toprule
Dataset    & \#Graphs & Avg.\# Nodes                                        & Avg.\# Edges                                        & \# Classes \\ \midrule
Cora       & 1        & 2,708                                               & 5,429                                               & 7          \\
PubMed     & 1        & 19,717                                              & 44,338                                              & 3          \\
ogbn-arXiv & 1        & 169,343                                             & 1,166,243                                           & 40         \\
Citeseer   & 1        & 3,312                                               & 4,732                                               & 6          \\
Wiki-CS    & 1        & 11,701                                              & 216,123                                             & 10         \\
BACE       & 1,513    & 34.1                                                & 73.7                                                & 1          \\
BBBP       & 2,039    & 23.9                                                & 51.6                                                & 1          \\
ClinTox    & 1,491    & 26.1 & 55.5 & 2          \\ \bottomrule
\end{tabular}
\caption{
Datasets statistics.
}
\label{tab:datasets}
\end{table}

\section{Baselines}
\label{baselines}
% \noindent\textbf{Baselines.} 
To evaluate the effectiveness of \ours, we compare it with the SOTA methods.
Details of these baselines are summarized as follows.

\noindent\textbf{{(1) MLP:}} Multilayer Perceptron (MLP) is a type of artificial neural network that consists of multiple layers of nodes (neurons) connected by weights and is primarily used for supervised learning tasks, such as classification.

\noindent\textbf{(2) Graph Neural Networks:}
\textbf{{GCN}}~\cite{kipf2016semi} is a fundamental method based on convolutional neural networks which operates directly on graph-structured data.
\textbf{{GAT}}~\cite{velivckovic2017graph} 
% based on graph convolutions 
computes the hidden representations of each node in the graph by first learning the importance of its neighbors and then aggregating information from them.
\textbf{{GIN}}~\cite{xu2019powerfulgraphneuralnetworks} develop a simple architecture that is provably the most expressive among the class of GNNs and is as powerful as the Weisfeiler-Lehman graph isomorphism test. 
\textbf{{GraphSAGE}}~\cite{hamilton2017inductive} present a general inductive framework that leverages node feature information (e.g., text attributes) to efficiently generate node embeddings for previously unseen data. 
\textbf{{Graphormer}}~\cite{ying2021transformers} is built upon the standard Transformer~\cite{vaswani2017attention} architecture, and could attain excellent results on a broad range of graph representation learning tasks. 
Graphormer propose several simple yet effective structural encoding methods to help Graphormer better model graph-structured data.
% \textbf{{GT}}~\cite{dwivedi2012generalization} presented a simple yet effective approach to generalize transformer networks on arbitrary graphs.
% It includes an elegant way to fuse node positional features using Laplacian eigenvectors for graph datasets, inspired from the heavy usage of positional encodings in NLP transformer models and recent research on node positional features in GNNs. 
% \textbf{{CoarFormer}}~\cite{kuang2024transformer} is a two-view architecture that captures fine-grained local information using a GNN-based module on the original graph and coarse yet long-range information using a Transformer-based module on the coarse graph.

{\noindent\textbf{(3) Ensemble Models:}
\textbf{{Bagging}}~\cite{breiman1996bagging} is an ensemble method that improves model stability and accuracy by training multiple base learners on bootstrap samples and aggregating their predictions through voting or averaging. 
\textbf{{Stacking}}~\cite{wolpert1992stacked} is an ensemble technique that combines predictions from multiple base learners to train a meta-learner, enhancing overall predictive performance through stacked generalization.
\textbf{{AdaBoost}}~\cite{freund1997decision} is an adaptive boosting algorithm that sequentially trains weak learners, adjusts sample weights, and combines their predictions to create a strong classifier with improved accuracy.}

\noindent\textbf{(4) LM Based Models:}
\textbf{{BERT}}~\cite{devlin2018bert} is a groundbreaking model in natural language processing.
It utilizes the transformer architecture to understand the context of words in a sentence by looking at both their left and right contexts simultaneously. This bidirectional approach enables BERT to capture nuanced meanings and relationships in text.
\textbf{{SentenceBERT}}~\cite{reimers2019sentence} is a modification of the original BERT (Bidirectional Encoder Representations from Transformers) model designed specifically for generating sentence embeddings. It was introduced to improve the performance of BERT on tasks that require understanding the semantic meaning of entire sentences, such as semantic textual similarity, paraphrase identification, and clustering.
\textbf{{DeBERTa}}~\cite{he2020deberta}
is a transformer-based language model that improves upon BERT and other models like RoBERTa by introducing several innovations. 
\textbf{{SciBERT}}~\cite{beltagy-etal-2019-scibert} is a BERT-based language model pre-trained on a large multi-domain corpus of scientific publications to achieve state-of-the-art performance on various scientific NLP tasks.
\textbf{{MedBERT}}~\cite{medbert} is a series of BERT and ALBERT models pre-trained on a large Chinese clinical corpus, which outperforms baseline models on named entity recognition and text classification tasks within the Chinese clinical NLP domain.

\begin{figure*}[h]
    \centering
    \includegraphics[width=0.85\linewidth]{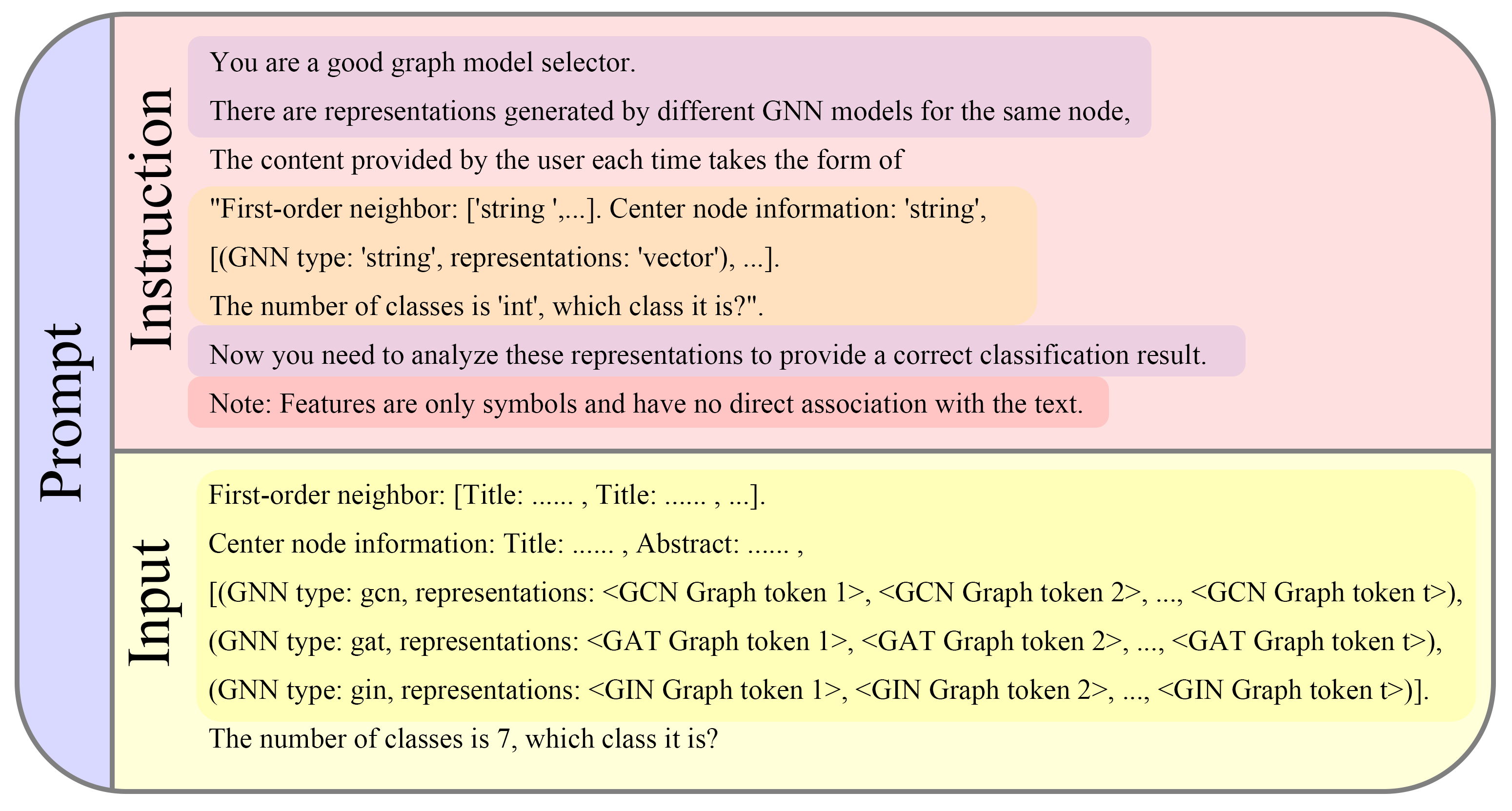}
    % \vspace{-1em}
    \caption{A prompt template for node classification task. }
    \label{fig:prompt-design-1}
\end{figure*}

\begin{figure*}[h]
    \centering
    \includegraphics[width=0.85\linewidth]{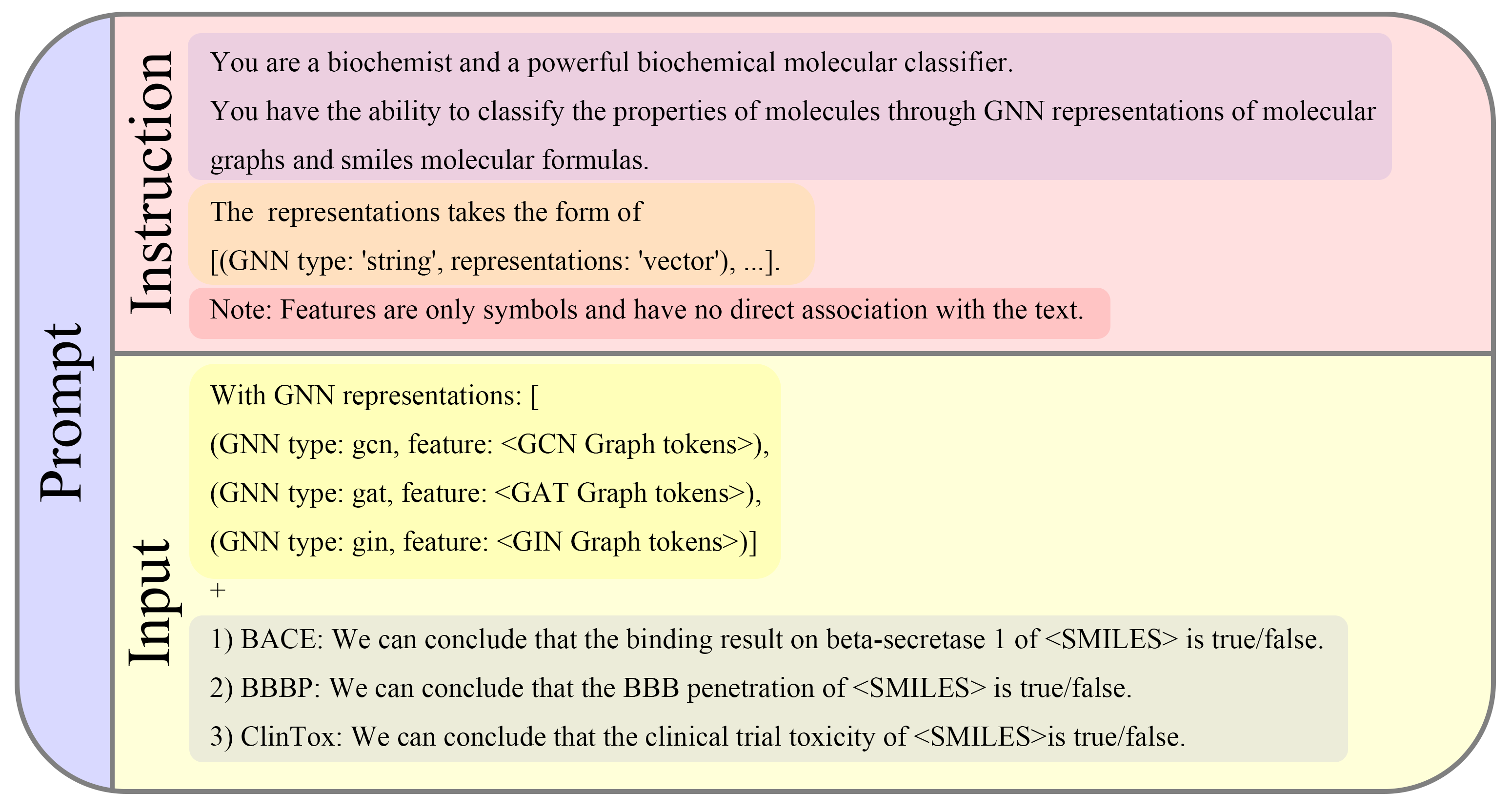}
    % \vspace{-1em}
    \caption{A prompt template for graph classification task. }
    \label{fig:prompt-design-2}
\end{figure*}

\noindent\textbf{(5) LLM Based Models:}
\textbf{{GraphGPT}}~\cite{tang2024graphgptgraphinstructiontuning} is a novel framework that integrates Large Language Models (LLMs) with graph structural knowledge through graph instruction tuning. This innovative approach enables LLMs to comprehend and interpret the structural components of graphs, thereby demonstrating superior generalization in both supervised and zero-shot graph learning tasks. \dhq{Additionally, GraphGPT employs Baichuan-7B and Vicuna-7B as its backbone models.}
\textbf{{LLaGA}}~\cite{chen2024llagalargelanguagegraph} integrates the capabilities of LLMs with graph-structured data, enabling LLMs to handle complex graph tasks effectively. LLaGA achieves this by reorganizing graph nodes into structure-aware sequences and mapping them into the token embedding space through a versatile projector, demonstrating superior versatility, generalizability, and interpretability across various datasets and tasks. \dhq{LLaGA employs Vicuna-7B as its backbone model.}
\textbf{{InstructGLM}}~\cite{ye2024language} use natural language to describe multi-scale geometric structure of the graph and then instruction finetune an LLM to perform graph tasks, which enables Generative Graph Learning. \dhq{InstructGLM employs Llama2-7B as its backbone model.}
\textbf{{DGTL}}~\cite{qin2024disentangledrepresentationlearninglarge}
\dhq{is an LLM-first-GNN-later method, which means first using LLMs to process text before training GNNs. DGTL }
incorporates graph structure information through tailored disentangled graph neural network layers, enabling LLMs to capture the intricate relationships hidden in TAGs from multiple structural factors,
which is able to enhance the reasoning and predicting capabilities of LLMs for TAGs.
\dhq{DGTL employs Llama2-13B as its backbone model.}
\textbf{{SNS-GPT4}}~\cite{li2024similaritybasedneighborselectiongraph} is a specialized version of the GPT-4 model designed for Social Network Services (SNS) applications. It introduces similarity-based neighbor selection to effectively improve the quality of selected neighbors, thereby enhancing graph representation and alleviating issues like over-squashing and heterophily.
\dhq{SNS-GPT4 employs GPT3.5-turbo and GPT4 as its backbone models.}
\dhq{Notably, SNS-GPT4 demonstrates better performance than TAPE~\cite{he2023explanations} (another LLM-first-GNN-later approach) in its reported results.}
% \textbf{{InstructGraph-INS}}~\cite{wang2024instructgraphboostinglargelanguage}
% is an advanced framework that integrates instruction-based learning with graph neural networks (GNNs) to improve performance in various graph-based tasks. The model leverages both explicit instructions and implicit graph structures to enhance representation learning and decision-making processes.
\textbf{{GAugLLM}}~\cite{fang2024gaugllmimprovinggraphcontrastive} is a novel framework for augmenting TAGs. It leverages advanced large language models like Mistral to enhance self-supervised graph learning. 
\dhq{GAugLLM employs Mistral 8*7b, ChatGPT-3.5 and
Llama2-13b as its backbone models.}
% \noindent\textbf{(5) 
\textbf{{Baichuan2}}~\cite{yang2023baichuan2openlargescale}
refers to the second iteration of the Baichuan large language model (LLM).
It has been trained on an impressive 2.6 trillion high-quality tokens, ensuring a robust understanding of language nuances.
Baichuan 2 comes in two main versions: 7B and 13B, both available in Base and Chat configurations, with the latter offering 4-bit quantization for efficient deployment.
\textbf{{OFA}}~\cite{liu2024alltraininggraphmodel} is a pioneering framework that unifies various graph tasks into a single model, enabling it to address classification tasks across different domains and tasks. OFA achieves this by converting graph data into text-attributed graphs (TAGs), using language models to encode diverse text attributes into a common embedding space, and introducing a novel graph prompting paradigm for in-context learning without fine-tuning. \dhq{OFA employs Llama2 as its backbone model.}
\textbf{{GOFA}}~\cite{kong2024gofagenerativeoneforallmodel} is a novel graph foundation model that integrates the strengths of large language models (LLMs) and graph neural networks (GNNs) to enable joint graph and language modeling. It achieves this by interleaving GNN layers into a frozen pre-trained LLM, allowing for self-supervised pretraining on graph data and fluidity in handling various graph-related tasks. \dhq{OFA employs Llama2 and Mistral as its backbone models.}

\section{Experiment setup}
\label{app:setup}
We implement \ours\ by PyTorch. 
We {primarily} use Baichuan2~\cite{yang2023baichuan2openlargescale} and InternLM2.5~\cite{cai2024internlm2} as the backbone LLM for \ours.
In the experiments,
we ensemble
% \ours\ ensembles 
% multiple 
three widely adopted GNN models: GCN, GAT and GIN, each with two layers. 
% Based on the validation set results, we conducted a grid search for the hyperparameters used in model fin-tuning. 
We perform grid search to fine-tune hyperparameters based on the validation set. 
Details on the search space is given in Table~\ref{tab:hyperparameters}.
We utilize LoRA+~\cite{hayou2024lora+} for fine-tuning.
Some fixed LoRA settings include capping each training sample at 2,047 tokens and using half-precision (FP16) for LoRA fine-tuning, with a batch size of 4 per GPU and gradient updates every step. 
We utilize a cosine-type learning rate scheduler and set the warmup ratio to 0.1. 
For the training of LLMs, we utilize Llama-Factory~\cite{zheng2024llamafactory} as a framework.
For GNN models, 
% following [xxx],
we use node representations obtained from the pre-trained {SentenceBERT}~\cite{reimers2019sentence} as input.
For baselines that report results on the adopted datasets, 
we directly report the results from their original papers.
For those whose results are missing,
we leave them blank.
% Due to computational power limitations, for those datasets where the original papers of the baselines did not report results, we left them blank in the table. 
For Baichuan2, 
we directly use it without fine-tuning.
We run all the experiments on a server with a single NVIDIA Tesla A100 GPU. 
% Additionally, due to the high computational cost of training LLM, the results we report are all from single runs.

{In addition, in the era of LLMs, due to the high cost, it is difficult to run experiments for baselines. For fair comparison, a widely adopted approach is to make comparison on the results for baselines reported from their original papers because their authors have well fine-tuned these models. This explains why only part of results are included in Table \ref{tab:performance-A} and why we did not reimplement some baselines. }

\begin{table}[!hbpt]
\centering
% \small
% \vspace{-1em}

\resizebox{1.0\linewidth}{!}{
\begin{tabular}{cc}
\toprule
\textbf{Hyperparameter} & \textbf{Search space} 
\\ \hline
learning rate           & $\{2.0 \times 10^{-5}, 5.0 \times 10^{-5}, 1.0 \times 10^{-4}\}$ \\
dropout                 & $[0,0.25]$ \\
loraplus lr ratio       & $\{16.0, 24.0, 32.0\}$ \\
Graph token $t$         & $\{1,2,4,8,16\}$ \\
\bottomrule
\end{tabular}
}
\caption{Hyperparameter Search Spaces.}
\label{tab:hyperparameters}
\end{table}

\section{Prompt design}
\label{prompt-details}

The prompt design follows three key principles:

\begin{enumerate}
    \item \textbf{Simultaneous input of textual strings from target node and its adjacent neighbors, graph tokens of the node, and task specification.} 
    Existing studies~\cite{kipf2016semi,velivckovic2017graph,verma2023bet} have demonstrated that aggregating the information from both the center node and its adjacent neighbors could contribute to the label prediction. Further,
    clear task instructions are necessary for accurate prediction.
    \item \textbf{ Differentiation between text tokens and graph tokens.} It directs LLM to correctly distinguish between textual tokens and graph tokens within different segments of the prompt, thus preventing confusion.
    \item \textbf{Guidance for learning from multi-GNNs:} It 
    leads LLM to implicitly learn how  
    to effectively combine strengths of different GNNs.
\end{enumerate}

The prompt is divided into two parts: Instruction and Input. The former specifies what the LLM should do with the input, defines the format of the input, and highlights the characteristics of the input content. The latter provides the specific text of target node or graph and their multiple GNN representations.

An example of the prompt template can be seen in Figure~\ref{fig:prompt-design-1}.
For graph classification, 
we follow MolXPT~\cite{liu2023molxptwrappingmoleculestext}
to design prompts for molecular graphs as shown in Figure~\ref{fig:prompt-design-2}. 
% \subsection{Node Classification}

\begin{figure}[!hbpt]
    \centering
    \includegraphics[width=0.95\linewidth]{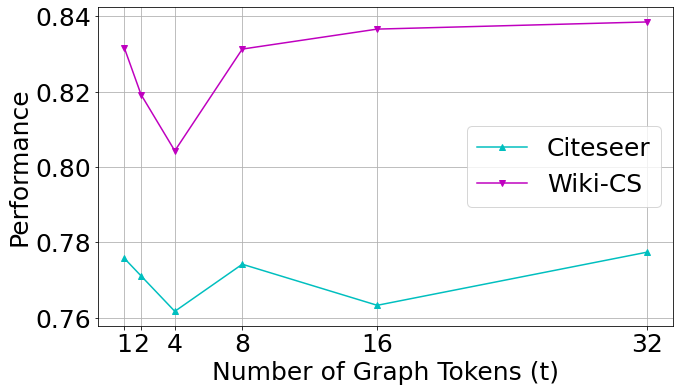}
    % \vspace{-1em}
    \caption{Hyperparameter sensitivity analysis.}
    % (Number of Graph Tokens $t$). 
    % \Description{The figure demonstrates that the performance initially increases and then gradually decreases as $t$ grows, with the optimal performance achieved at $t = 8$.}
    % }
    \label{fig:hpa}
\end{figure}

{\section{Best variants}
\label{app:best-variants}
The best variants in Table \ref{tab:performance-A} and Table \ref{tab:performance-20shot} are listed in Table \ref{tab:variants-A} and Table \ref{tab:variants-20shot}, respectively. 

\begin{table}
        \centering
        \resizebox{1.0\linewidth}{!}{
            \begin{tabular}{c|ccc}
            \toprule
                \textbf{Backbone LLM} & \textbf{Cora} & \textbf{PubMed} & \textbf{ogbn-arXiv} \\
        \hline

        InternLM2.5-7B & \ours-ALL & \ours-ALL & \ours-[GAT+GIN] \\
        Baichuan2-7B & \ours-[GCN+GAT] & \ours-ALL & \ours-ALL \\
        Baichuan2-13B & \ours-[GCN+GIN] & \ours-ALL & \ours-ALL \\

            \bottomrule
            \end{tabular}
            }
        \caption{Best variants in Table  \ref{tab:performance-A}.}
        \label{tab:variants-A}
\end{table}

\begin{table}
        \centering
        \resizebox{0.85\linewidth}{!}{
            \begin{tabular}{c|ccc}
            \toprule
            \textbf{Backbone LLM} & \textbf{Cora} & \textbf{PubMed} \\
            \hline
            Baichuan2-7B & \ours-ALL & \ours-ALL \\
            InternLM2.5-7B & \ours-ALL & \ours-[GAT+GIN] \\
            \bottomrule
            \end{tabular}
            }
        \caption{Best variants in Table \ref{tab:performance-20shot}.}
        \label{tab:variants-20shot}
\end{table}

\section{LLama2-7B as backbone}
\label{app:llama2}
We have supplemented our experiments by implementing LLaMA2-7B as the backbone LLM. We conducted these experiments on a medium-scale dataset, PubMed, as a representative, with results presented in Table~\ref{tab:llama2}:

\begin{table}[!hbpt]
        \centering
        \resizebox{0.7\linewidth}{!}{
        \begin{tabular}{l|ccccc}
        \toprule
        \textbf{Model} & \textbf{PubMed} \\
        \hline

% \textit{Traditional GNNs} \\
MLP & 82.40 \\
GCN & 83.78 \\
GAT & 83.59 \\
GIN & 82.03 \\
GraphSAGE & 87.01 \\
Graphormer & 88.75 \\
\midrule
% \textit{Language Models} \\
BERT & 93.91 \\
SentenceBERT & 92.49 \\
DeBERTa & 93.45 \\
\midrule
% \textit{Graph-LLM Methods} \\
GraphGPT-MIX-7B & 64.76 \\
LLaGA-HO-7B(GENERAL) & 75.01 \\
InstructGLM-Llama-7B & 93.84 \\
DGTL & 87.10 \\
SNS-GPT4 & 93.80 \\
GAugLLM & 74.15 \\
Baichuan2-13B & 36.04 \\
\midrule
\textbf{LensGNN*-LLaMA2-7B} & \textbf{94.59} \\

        \bottomrule
    \end{tabular}
        }
        \caption{Performance comparison (\%) on PubMed dataset using LLaMA2-7B as backbone. (\%)}
        \label{tab:llama2}
\end{table}

Experimental results demonstrate that our model with LLaMA2-7B as the backbone also achieves superior accuracy compared to: (1) GNN baselines, (2) LM-based methods, and (3) other LLM-based approaches.

\section{Additional hyperparameter sensitivity analysis}
\label{app:add-hyperparameter-analysis}
We further conduct hyperparameter sensitivity analysis on Citeseer and Wiki-CS,
whose results are given 
% As shown in the results 
in Figure \ref{fig:hpa}.
From the figure, we see that
the model can also achieve good results at $t=8$ in terms of the overall performance, which is similar as the results in Figure~\ref{fig:hp}.}

\section{One-shot results}
\label{app:one-shot}
We further conduct experiments on one-shot setting, the results are presented in Table \ref{tab:one-shot}.
From the table, we see that
\begin{table}
        \centering
        \resizebox{0.85\linewidth}{!}{
        \begin{tabular}{c|cccc}
        \toprule
        \textbf{Model} & \textbf{Cora} & \textbf{PubMed} & \textbf{Citeseer} & \textbf{Wiki-CS} \\
        \hline
GCN	& 18.90	& 41.55	& 32.42	& 28.73 \\
GAT	& 25.43	& 49.93	& 35.06	& \textbf{40.94} \\
GIN	& 20.21	& 40.71	& 27.63	& 32.78 \\
        \hline
        \ours & \textbf{59.38} & \textbf{83.79} & \textbf{53.32} & 30.47 \\
        \bottomrule
    \end{tabular}
        }
        \caption{Classification accuracy in one-shot setting. (\%)}
        \label{tab:one-shot}
\end{table}
our method still achieves outstanding performance on the Cora, PubMed, and Citeseer datasets, which shows the strong generalizability. For WikiCS, due to the very scare labeled data, GNN ensembling is adversely affected.

\section{Input length analysis}
\label{app-input-length}

\begin{figure*}[h]
    \centering
    \includegraphics[width=1.0\linewidth]{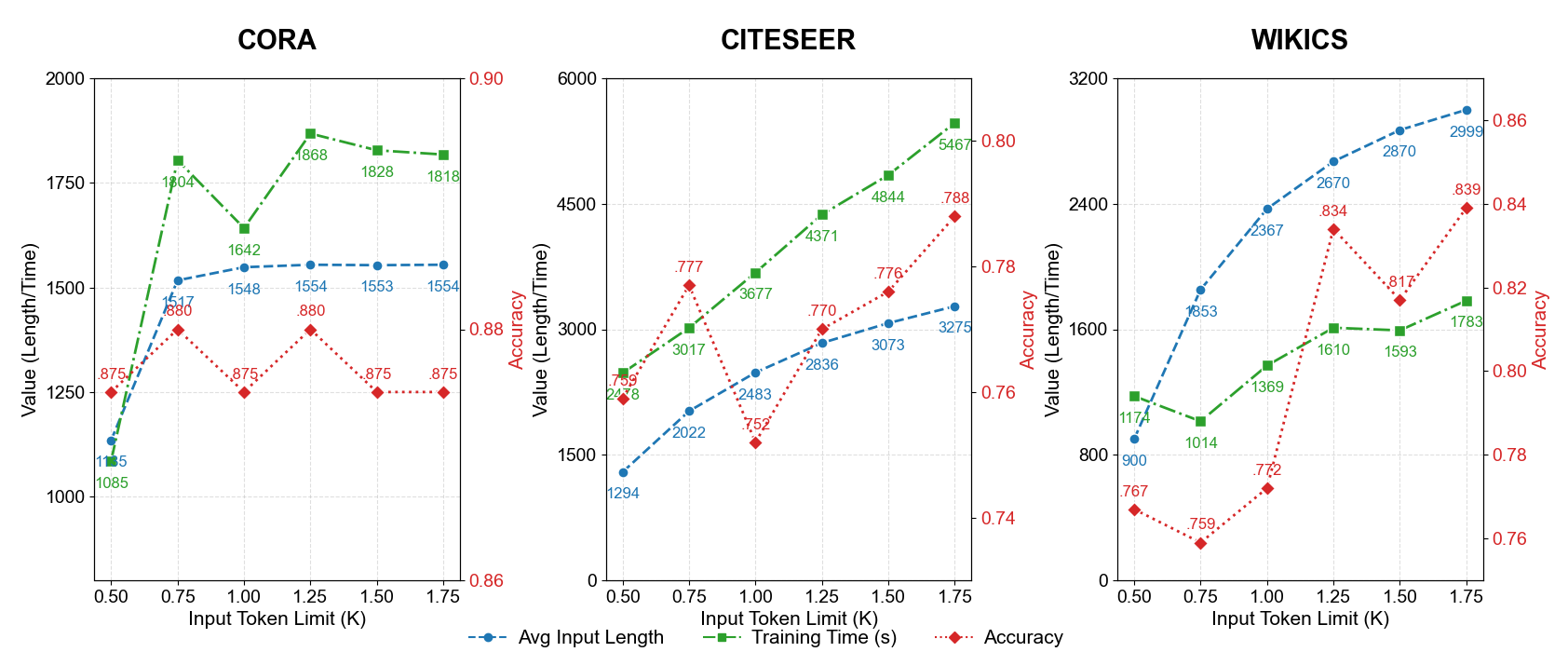}
    % \vspace{-1em}
    \caption{Input length analysis. }
    \label{fig:input_length}
\end{figure*}

We investigate the impact of LLM input length on model efficiency and performance across Cora, Citeseer, and Wiki-CS. Specifically, we established different limits on total number of input tokens. When exceeded, some neighbor text is randomly truncated. Within this constraint, more text introduces additional noise but also provides richer useful information, leading to better model performance—albeit at increased computational cost. The experimental results are presented in Figure~\ref{fig:input_length}.

Our experiments using Baichuan-7B as the backbone record sample statistics with total token length limits ranging from 0.5K to 1.75K, including average input string length, training time, and test accuracy. While Cora's shorter texts show negligible variations, both Citeseer and Wiki-CS exhibit a clear trend: longer inputs increased training time but also improved accuracy.

\section{Detailed hyperparameter settings}
\label{app-hyperparameters}

% The hyperparameter search space is shown in the Table~\ref{tab:hyperparameters}. 
The details on the setting of hyperparameters in the first and second stages of \ours~are shown in Tables \ref{tab:detailed_hyperparameters_1} and \ref{tab:detailed_hyperparameters_2}, respectively. 
The hyperparameters for the 20-shot experiment are shown in Tables \ref{tab:detailed_hyperparameters_1_20shot} and \ref{tab:detailed_hyperparameters_2_20shot}.

\begin{table*}[!hbpt]
% \vspace{-1em}
\centering

\resizebox{1.0\linewidth}{!}{
\begin{tabular}{c|cccccccc}
\toprule
Model & Dataset & Hidden size & \makecell{GNN representation\\size}  & \makecell{Number of\\layers} & \makecell{Learning rate of\\GNN encoder} & \makecell{Initial learning rate of\\the classifier} & Dropout Rate & Weight Decay \\
\hline
\multirow{8}{*}{\textbf{GCN}}
& Cora & 256 & 40960 & 2 & 0.0002 & 0.0002 & 0.25 & 0.001 \\
& Pubmed & 256 & 40960 & 2 & 0.0002 & 0.0002 & 0.25 & 0.001 \\
& ogbn-arxiv & 256 & 20480 & 2 & 0.01 & 0.0005 & 0.2 & 0 \\
& Citeseer & 256 & 40960 & 2 & 0.0001 & 0.0005 & 0.25 & 0.001 \\
& Wiki-CS & 256 & 40960 & 2 & 0.0002 & 0.0005 & 0.25 & 0.001 \\
& BACE & 256 & 40960 & 4 & 0.0001 & 0.0005 & 0.1 & 0.0005 \\
& BBBP & 256 & 40960 & 2 & 0.0001 & 0.0005 & 0.1 & 0.0005 \\
& ClinTox & 256 & 40960 & 2 & 0.0001 & 0.0005 & 0.1 & 0.0005\\
\hline
\multirow{8}{*}{\textbf{GAT}}
& Cora & 256 & 40960 & 2 & 0.0002 & 0.0002 & 0.25 & 0.001 \\
& Pubmed & 256 & 40960 & 2 & 0.0002 & 0.0002 & 0.25 & 0.001 \\
& ogbn-arxiv & 256 & 20480 & 2 & 0.01 & 0.0005 & 0.2 & 0 \\
& Citeseer & 256 & 40960 & 2 & 0.0001 & 0.0005 & 0.25 & 0.001 \\
& Wiki-CS & 256 & 40960 & 2 & 0.0002 & 0.0005 & 0.25 & 0.001 \\
& BACE & 256 & 40960 & 4 & 0.0001 & 0.0005 & 0.1 & 0.0005 \\
& BBBP & 256 & 40960 & 2 & 0.0001 & 0.0005 & 0.1 & 0.0005 \\
& ClinTox & 256 & 40960 & 2 & 0.0001 & 0.0005 & 0.1 & 0.0005\\
\hline
\multirow{8}{*}{\textbf{GIN}}
& Cora & 256 & 40960 & 2 & 0.0001 & 0.0001 & 0.25 & 0.001 \\
& Pubmed & 256 & 40960 & 2 & 0.0001 & 0.0001 & 0.25 & 0.001 \\
& ogbn-arxiv & 256 & 20480 & 2 & 0.004 & 0.0005 & 0.2 & 0 \\
& Citeseer & 256 & 40960 & 2 & 0.00005 & 0.0005 & 0.25 & 0.001 \\
& Wiki-CS & 256 & 40960 & 2 & 0.0001 & 0.0005 & 0.25 & 0.001 \\
& BACE & 256 & 40960 & 4 & 0.0001 & 0.0005 & 0.1 & 0.0005 \\
& BBBP & 256 & 40960 & 2 & 0.0001 & 0.0005 & 0.1 & 0.0005 \\
& ClinTox & 256 & 40960 & 2 & 0.0001 & 0.0005 & 0.1 & 0.0005\\
\bottomrule
\end{tabular}
}
\caption{Detailed hyperparameters of aligning multi-GNNs.}

\label{tab:detailed_hyperparameters_1}
\end{table*}

\begin{table*}[!hbpt]
\centering
% \vspace{-1em}
\resizebox{1.0\linewidth}{!}{
\begin{tabular}{c|ccccccccc} 
\toprule 
Model & Dataset & \makecell{Lora dropout} & \makecell{Loraplus\\lr ratio} & \makecell{Training\\batch size} & \makecell{Learning rate} & \makecell{Early stop\\epoch} & \makecell{Warmup ratio} \\ 
\hline 
\multirow{8}{*}{Baichuan2-13B}
& Cora & 0.1 & 16 & 4 & 0.0001 & 3 & 0.1 \\ 
& Pubmed & 0.1 & 16 & 4 & 0.00005 & 2 & 0.1 \\ 
& ogbn-arxiv & 0.1 & 16 & 4 & 0.0001 & 3 & 0.1 \\ 
& Citeseer & 0.15 & 16 & 4 & 0.0001 & 5 & 0.1 \\
& Wiki-CS & 0.1 & 16 & 4 & 0.0001 & 5 & 0.1 \\
& BACE & 0.1 & 16 & 4 & 0.0001 & 4 & 0.1 \\
& BBBP & 0.1 & 16 & 4 & 0.0001 & 3 & 0.1 \\
& ClinTox & 0.1 & 16 & 4 & 0.0001 & 3 & 0.1 \\
\hline 
\multirow{3}{*}{Baichuan2-7B}
& Cora & 0.15 & 16 & 4 & 0.0001 & 3 & 0.1 \\
& Pubmed & 0.15 & 16 & 4 & 0.0001 & 2 & 0.1 \\
& ogbn-arxiv & 0.15 & 16 & 4 & 0.0001 & 3 & 0.1 \\
\hline 
\multirow{3}{*}{InternLM2.5-7B}
& Cora & 0.1 & 16 & 4 & 0.0001 & 3 & 0.1 \\ 
& Pubmed & 0.1 & 16 & 4 & 0.0001 & 2 & 0.1 \\ 
& ogbn-arxiv & 0.1 & 16 & 4 & 0.0001 & 3 & 0.1 \\
\hline 
\multirow{2}{*}{Falcon-7B}
& Cora & 0.1 & 16 & 4 & 0.0001 & 3 & 0.1 \\ 
& Pubmed & 0.1 & 16 & 4  & 0.0001 & 2 & 0.1 \\ 
\bottomrule \end{tabular}
}
\caption{Detailed hyperparameters of Ensembling multi-GNNs with LLM (LoRA fine-tune).}
\label{tab:detailed_hyperparameters_2}
\end{table*}

\begin{table*}[!hbpt]
% \vspace{-1em}
\centering

\resizebox{1.0\linewidth}{!}{
\begin{tabular}{c|cccccccc}
\toprule
Model & Dataset & Hidden size & \makecell{GNN representation\\size}  & \makecell{Number of\\layers} & \makecell{Learning rate of\\GNN encoder} & \makecell{Initial learning rate of\\the classifier} & Dropout Rate & Weight Decay \\
\hline
\multirow{2}{*}{\textbf{GCN}}
& Cora & 256 & 32768 & 2 & 0.0002 & 0.0005 & 0.25 & 0.001 \\
& Pubmed & 256 & 32768 & 2 & 0.0002 & 0.0005 & 0.25 & 0.001 \\
\hline
\multirow{2}{*}{\textbf{GAT}}
& Cora & 256 & 32768 & 2 & 0.0002 & 0.0005 & 0.25 & 0.001 \\
& Pubmed & 256 & 32768 & 2 & 0.0002 & 0.0005 & 0.25 & 0.001 \\
\hline
\multirow{2}{*}{\textbf{GIN}}
& Cora & 256 & 32768 & 2 & 0.0001 & 0.0005 & 0.25 & 0.001 \\
& Pubmed & 256 & 32768 & 2 & 0.0001 & 0.0005 & 0.25 & 0.001 \\
\bottomrule
\end{tabular}
}
\caption{Detailed hyperparameters of aligning multi-GNNs (20-shot).}

\label{tab:detailed_hyperparameters_1_20shot}
\end{table*}

\begin{table*}[h]
\centering
% \vspace{-1em}
\resizebox{1.0\linewidth}{!}{
\begin{tabular}{c|ccccccccc} 
\toprule 
Model & Dataset & \makecell{Lora dropout} & \makecell{Loraplus\\lr ratio} & \makecell{Training\\batch size} & \makecell{Learning rate} & \makecell{Early stop\\epoch} & \makecell{Warmup ratio} \\ 
\hline 
\multirow{2}{*}{Baichuan2-7B}
& Cora & 0.15 & 16 & 4 & 0.0001 & 8 & 0.1 \\
& Pubmed & 0.15 & 16 & 4 & 0.0001 & 9 & 0.1 \\
\hline 
\multirow{2}{*}{InternLM2.5-7B}
& Cora & 0.1 & 16 & 4 & 0.0001 & 8 & 0.1 \\ 
& Pubmed & 0.05 & 16 & 4 & 0.0001 & 9 & 0.1 \\ 
\bottomrule \end{tabular}
}
\caption{Detailed hyperparameters of Ensembling multi-GNNs with LLM (LoRA fine-tune) (20-shot).}
\label{tab:detailed_hyperparameters_2_20shot}
\end{table*}

% \begin{table*}[h]
% \centering
% \caption{Detailed hyperparameters of Ensembling multi-GNNs with LLM (LoRA fine-tune).}
% % \vspace{-1em}
% \resizebox{1.0\linewidth}{!}{
% \begin{tabular}{c|ccc|cc|cc}
% \toprule
% Model & \multicolumn{3}{c|}{Baichuan2-13B} & \multicolumn{2}{c|}{Falcon-7B} & \multicolumn{2}{c}{InternLM2.5-7B-chat} \\
% \hline
% Dataset & Cora & Pubmed & ogbn-arxiv & Cora & Pubmed & Cora & Pubmed \\
% \hline
% Lora dropout & 0.1 & 0.1 & 0.1 & 0.1 & 0.1 & 0.1 & 0.1 \\
% Loraplus Ir ratio & 16 & 16 & 16 & 16 & 16 & 16 & 16 \\
% Prompt cutoff length & 2047 & 2047 & 2047 & 2047 & 2047 & 2047 & 2047 \\
% Training batch size per device & 4 & 4 & 4 & 4 & 4 & 4 & 4 \\
% Gradient accumulation steps & 1 & 1 & 1 & 1 & 1 & 1 & 1 \\
% Learning rate & 0.0001 & 0.00005 & 0.0001 & 0.0001 & 0.0001 & 0.0001 & 0.0001 \\
% Early stop epoch & 3 & 2 & 3 & 3 & 2 & 3 & 2 \\
% Warmup Ratio & 0.1 & 0.1 & 0.1 & 0.1 & 0.1 & 0.1 & 0.1 \\
% Parameter precision & FP16 & FP16 & FP16 & FP16 & FP16 & FP16 & FP16 \\
% \bottomrule
% \end{tabular}
% }
% \label{tab:detailed_hyperparameters_2}
% \end{table*}

\end{document}